\documentclass{article}

\usepackage[final, nonatbib]{neurips_2019}

\usepackage[utf8]{inputenc} 
\usepackage[T1]{fontenc}    
\usepackage{hyperref}       
\usepackage{url}            
\usepackage{booktabs}       
\usepackage{amsfonts}       
\usepackage{nicefrac}       
\usepackage{microtype}      

\usepackage{subcaption}
\usepackage{caption}
\usepackage{multirow}
\usepackage{amsmath}
\usepackage{amsthm}
\usepackage{amssymb}
\usepackage{tikz}

\DeclareMathOperator*{\argmin}{arg\,min}
\usepackage{enumitem}

\newcommand{\E}{\mathbb{E}}

\newcommand{\one}{\mathbb{I}}

\usepackage{graphicx}
\usepackage{amsmath}
\newcommand{\distas}[1]{\mathbin{\overset{#1}{\kern\z@\sim}}}%
\usepackage{algorithm}
\usepackage{algorithmic}

\usepackage{wrapfig}
\usepackage{lipsum}

\usepackage[textsize=scriptsize]{todonotes}

\usepackage{xcolor}
\usetikzlibrary{decorations.pathreplacing}

\definecolor{tomato}{HTML}{FF0000}
\definecolor{dodgerblue}{HTML}{0000FF}
\definecolor{orange}{HTML}{00FF00}
\definecolor{lime}{HTML}{FFA500}

\usepackage{amsmath,amsfonts,bm}

\def\eqref#1{equation~\ref{#1}}

\def\1{\bm{1}}

\def\vx{{\bm{x}}}

\def\vz{{\bm{z}}}
\def\veps{{\bm{\epsilon}}}
\def\vdel{{\bm{\delta}}}

\DeclareMathAlphabet{\mathsfit}{\encodingdefault}{\sfdefault}{m}{sl}
\SetMathAlphabet{\mathsfit}{bold}{\encodingdefault}{\sfdefault}{bx}{n}

\def\gB{{\mathcal{B}}}

\def\gD{{\mathcal{D}}}

\def\gF{{\mathcal{F}}}

\def\gL{{\mathcal{L}}}

\def\gN{{\mathcal{N}}}

\def\gX{{\mathcal{X}}}
\def\gY{{\mathcal{Y}}}

\def\sP{{\mathbb{P}}}

\def\sR{{\mathbb{R}}}

\def\sZ{{\mathbb{Z}}}

\usepackage{xcolor}

\usepackage{amsthm}
\newtheorem{theorem}{Theorem}
\newtheorem{lemma}[theorem]{Lemma}

\newtheorem{remark}[theorem]{Remark}

\newtheorem{proposition}[theorem]{Proposition}

\newcommand{\textred}[1]{\textcolor{black}{#1}}
\newcommand{\textcyan}[1]{\textcolor{black}{#1}}
\newcommand{\textblue}[1]{\textcolor{black}{#1}}
\newcommand{\textgrey}[1]{\textcolor{black}{#1}}

\newcommand{\xref}[1]{\S\ref{#1}}

\usepackage{mathtools}

\newcommand{\header}[1]{\vspace{0.05in}\noindent\textbf{#1}}
\newcommand{\rand}{\phi}
\newcommand{\baserand}{\varphi}

\title{
Tight Certificates of Adversarial Robustness\\
for Randomly Smoothed Classifiers
}

\author{Guang-He Lee$^1$, Yang Yuan$^{1,2}$, Shiyu Chang$^3$, Tommi S. Jaakkola$^1$\\
$^1$MIT Computer Science and Artificial Intelligence Lab \\
$^2$Institute for Interdisciplinary Information Sciences, Tsinghua University\\
$^3$MIT-IBM Watson AI Lab\\
{\tt\small \{guanghe, yangyuan, tommi\}@csail.mit.edu}, ~{\tt\small shiyu.chang@ibm.com}}

\begin{document}
\maketitle

\begin{abstract}
Strong theoretical guarantees of robustness can be given for ensembles of classifiers generated by input randomization. Specifically, an $\ell_2$ bounded adversary cannot alter the ensemble prediction generated by an additive isotropic Gaussian noise, where the radius for the adversary depends on both the variance of the distribution as well as the ensemble margin at the point of interest. We build on and considerably expand this work across broad classes of distributions. In particular, we offer adversarial robustness guarantees and associated algorithms for the discrete case where the adversary is $\ell_0$ bounded. Moreover, we exemplify how the guarantees can be tightened with specific assumptions about the function class of the classifier such as a decision tree. We empirically illustrate these results with and without functional restrictions across image and molecule datasets.\footnote{{Project page: \url{http://people.csail.mit.edu/guanghe/randomized_smoothing}}.}
\end{abstract}

\section{Introduction}

Many powerful classifiers lack robustness in the sense that a slight, potentially unnoticeable manipulation of the input features, e.g., by an adversary, can cause the classifier to change its prediction~\cite{goodfellow14}. The effect is clearly undesirable in decision critical applications. Indeed, a lot of recent work has gone into analyzing such failures together with providing certificates of robustness. 

Robustness can be defined with respect to a variety of metrics that bound the magnitude or the type of adversarial manipulation. The most common approach to searching for violations is by finding an adversarial example within a small neighborhood of the example in question, e.g., using gradient-based algorithms~\cite{finlay2019logbarrier, goodfellow14, madry2017towards}. The downside of such approaches is that failure to discover an adversarial example does not mean that another technique could not find one. For this reason, a recent line of work has instead focused on certificates of robustness, i.e., guarantees that ensure, for specific classes of methods, that no adversarial examples exist within a certified region. Unfortunately, obtaining exact guarantees can be computationally intractable~\cite{katz2017reluplex, lomuscio2017approach, tjeng2017evaluating}, and guarantees that scale to realistic architectures have remained somewhat conservative~\cite{croce2018provable, mirman2018differentiable, weng2018towards, wong2017provable, zhang2018efficient}.

Ensemble classifiers have recently been shown to yield strong guarantees of robustness~\cite{cohen2019certified}. The ensembles, in this case, are simply induced from randomly perturbing the input to a base classifier. The guarantees state that, given an \textgrey{additive isotropic Gaussian noise} on the input example, an adversary cannot alter the prediction of the corresponding ensemble within an $\ell_2$ radius, where the radius depends on the noise variance as well as the ensemble margin at the given point~\cite{cohen2019certified}. 

In this work, we substantially extend robustness certificates for such noise-induced ensembles. We provide guarantees for alternative metrics and noise distributions (e.g., uniform), develop a stratified likelihood ratio analysis that allows us to provide \textblue{certificates of robustness over discrete spaces with respect to $\ell_0$ distance, which are \emph{tight} and \emph{applicable} to any measurable classifiers.} We also introduce scalable algorithms for computing the certificates. The guarantees can be further tightened by introducing additional assumptions about the family of classifiers. We illustrate this in the context of ensembles derived from decision trees. Empirically, our ensemble classifiers yield \textblue{the state-of-the-art} certified guarantees with respect to $\ell_0$ bounded adversaries across image and molecule datasets in comparison to the previous methods adapted from continuous spaces. 

\section{Related Work}

In a classification setting, the role of \textgrey{robustness} certificates is to guarantee a constant classification within a local region; a certificate is always sufficient to claim robustness.  When a certificate is both sufficient and necessary, it is called an exact certificate.  For example, the exact $\ell_2$ certificate of a linear classifier is the $\ell_2$ distance between the classifier and a given point. Below we focus the discussions on the recent development of robustness guarantees for deep networks. 

Most of the exact methods are derived on piecewise linear networks, defined as any network architectures with piecewise linear activation functions. Such class of networks has a mix integer-linear representation~\cite{lee2018towards}, which allows the usage of mix integer-linear programming~\cite{cheng2017maximum, dutta2018output, fischetti2018deep, lomuscio2017approach, tjeng2017evaluating} or satisfiability modulo theories~\cite{carlini2017provably, ehlers2017formal, katz2017reluplex, scheibler2015towards} to find the exact adversary under an $\ell_q$ radius. However, the exact method is in general NP-complete, and thus does not scale to \textcyan{large} problems~\cite{tjeng2017evaluating}.

A certificate that only holds a sufficient condition is conservative but can be more scalable than exact methods. Such guarantees may be derived as a linear program~\cite{wong2017provable, wong2018scaling}, a semidefinite program~\cite{raghunathan2018certified, raghunathan2018semidefinite}, or a dual optimization problem~\cite{dvijotham2018training, dvijotham2018dual} through relaxation. Alternative approaches conduct layer-wise relaxations of feasible neuron values to derive the certificates~\cite{gowal2018effectiveness, mirman2018differentiable, singh2018fast, weng2018towards, zhang2018efficient}. Unfortunately, there is no empirical evidence of an effective certificate from the above methods in large scale problems. This does not entail that the certificates are not tight enough in practice; it might also be attributed to the fact that it is challenging to obtain a robust network in a large scale setting. 

Recent works propose a new modeling scheme that ensembles a classifier by input randomization~\cite{cao2017mitigating, liu2018towards}, mostly done via an \textgrey{additive isotropic Gaussian noise}. Lecuyer et al.~\cite{lecuyer2018certified} first propose a certificate based on differential privacy, which is improved by Li et al.~\cite{li2018second} using R$\acute{\text{e}}$nyi divergence. Cohen et al.~\cite{cohen2019certified} proceed with the analysis by proving the \emph{tight} certificate with respect to \emph{all the measurable classifiers} based on the Neyman-Pearson Lemma~\cite{neyman1933ix}, which yields the state-of-the-art provably robust classifier. However, the tight certificate is tailored to an isotropic Gaussian distribution and $\ell_2$ \textgrey{metric}, while we generalize the result across broad classes of distributions and metrics. In addition, we show that such tight guarantee can be tightened with assumptions about the classifier.

\textblue{Our method of certification also yields the first {tight} and actionable $\ell_0$ robustness certificates in discrete domains (cf. continuous domains where an adversary is easy to find~\cite{goodfellow14}). Robustness guarantees in discrete domains are combinatorial in nature and thus challenging to obtain. Indeed, even for simple binary vectors, verifying robustness requires checking an exponential number of predictions for any black-box model.\footnote{\textblue{We are aware of two concurrent works also yielding certificates in discrete domain~\cite{huang2019achieving, jia2019certified}.}}}

\section{\textblue{Certification Methodology}}
\label{sec:certificates}
\textcyan{
Given an input $\vx \in \gX$, a {randomization scheme} $\rand$ assigns a probability mass/density $\Pr(\rand(\vx) = \vz)$ for each randomized outcome $\vz \in \gX$. We can define a probabilistic classifier either by specifying the associated conditional distribution $\sP(y | \vx)$ for a class $y\in \gY$ or by viewing it as a random function $f(\vx)$ where the randomness in the output is independent for each $\vx$.  We compose the {randomization scheme} $\phi$ with a classifier $f$ to get a randomly smoothed classifier $\E_{\phi}[\sP(y | \phi(\vx))]$, where the probability for outputting a class $y \in \gY$ is denoted as $\Pr(f(\phi(\vx))=y)$ and abbreviated as $p$, whenever $f, \rand, \vx$ and $y$ are clear from the context.} 
\textcyan{Under this setting, we first develop our framework for tight robustness certificates in \xref{sec:framework}, exemplify the framework in \xref{sec:warmup}-\ref{sec:l0-perturb}, and illustrate how the guarantees can be refined with further assumption in \xref{sec:comparison}-\ref{sec:refinement}. We defer all the proofs to Appendix \ref{sec:proofs}.}

\subsection{A Framework for Tight Certificates of Robustness}
\label{sec:framework}

In this section, we develop our {framework} for deriving tight certificates of robustness for randomly smoothed classifiers, which will be {instantiated} in the following sections. 

\header{Point-wise Certificate.} Given $p$, we first identify a tight lower bound on the probability score $\Pr(f(\rand(\bar \vx)) = y)$ for another (neighboring) point $\bar \vx \in \gX$. Here we denote the set of measurable classifiers with respect to $\rand$ as $\gF$.  \textcyan{Without any additional assumptions on $f$,} a lower bound can be found by the minimization problem:
\begin{equation}
\rho_{\vx, \bar \vx}(p) \triangleq \min_{\bar{f} \in \gF: \Pr(\bar{f}(\rand(\vx)) = y) = p} \Pr(\bar{f}(\rand(\bar{\vx})) = y) \leq \Pr(f(\rand(\bar \vx)) = y). 
\label{eqn:target}    
\end{equation}
Note that bound is tight since $f$ satisfies the constraint. 

\header{Regional Certificate.} We can extend {the point-wise certificate} $\rho_{\vx, \bar \vx}(p)$ to \textcyan{a regional certificate} by examining the worst case $\bar \vx$ over the neighboring region around $\vx$.  Formally, given an $\ell_q$ metric $\|\cdot\|_q$, the neighborhood around $\vx$ with radius $r$ is defined as $\gB_{r, q}(\vx) \triangleq \{\bar \vx \in \gX: \| \vx - \bar \vx \|_q \leq r\}$. Assuming $p = \Pr(f(\rand(\vx)) = y) > 0.5$ for a $y \in \gY$, a robustness certificate on the $\ell_q$ radius can be found by
\begin{equation}
    R(\vx, p, q) \triangleq \sup r, \text{  }s.t. \min_{\bar{\vx} \in \gB_{r, q}(\vx)}
    \rho_{\vx, \bar \vx}(p)  > 0.5. 
    \label{eq:main_lemma}
\end{equation}
Essentially, the certificate $R(\vx, p, q)$ entails the following robustness guarantee:
\begin{equation}
    \forall \bar \vx \in \gX: \|\vx - \bar \vx\|_q < R(\vx, p, q),\text{we have } \Pr(f(\phi(\bar \vx)) = y) > 0.5.
\end{equation}
When the maximum can be attained in Eq.~(\ref{eq:main_lemma}) (which will be the case in $\ell_0$ norm), the above $<$ can be replaced with $\leq$.
Note that here we assume $\Pr(f(\rand(\vx)) = y) > 0.5$ and ignore the {case} that $0.5 \geq \Pr(f(\rand(\bar \vx)) = y) > \max_{y'\neq y} \Pr(f(\rand(\bar \vx)) = y')$. {By definition,} the certified radius $R(\vx, p, q)$ is tight for binary classification, and {provides} a reasonable sufficient condition to guarantee robustness for $|\gY| > 2$.  The tight guarantee for $|\gY| > 2$ will involve the maximum prediction probability over all the remaining classes (see Theorem 1 of \cite{cohen2019certified}).  However, when the prediction probability $p = \Pr(f(\rand(\vx)) = y)$ is intractable to compute and relies on statistical estimation for each class $y$ (e.g., when $f$ is a deep network), the tight guarantee is statistically challenging to obtain. {The actual algorithm used by Cohen et al.~\cite{cohen2019certified}} is also a special case of Eq.~(\ref{eq:main_lemma}).

\subsection{A Warm-up Example: the Uniform Distribution}
\label{sec:warmup}

\begin{wrapfigure}{R}{0.2\textwidth}
\vspace{-12mm}
    \centering
\begin{tikzpicture}
 \draw [fill=yellow, fill opacity=0.2]
      (0,0) -- (0,2) -- (2,2) -- (2,0)-- cycle;
 \draw [fill=red, fill opacity=0.2]
      (0.5,0.5) --+ (0,2) --+ (2,2) --+ (2,0)-- cycle;
\node at (0.2,2.3) {$\gL_4$};
\node at (0.3,0.5) {$\gL_1$};
\node at (2.3,1.7) {$\gL_3$};
\node at (1.1,1.4) {$\gL_2$};
\node[draw,shape=circle,fill=black,minimum size=1mm, inner sep=0.04mm,outer sep=0pt, 
label=below:$\vx$] at (1,1) {};
\node[draw,shape=circle,fill=black,minimum size=1mm, inner sep=0.04mm,outer sep=0pt, 
label=above:$\bar \vx $] at (1.5, 1.5) {};
\end{tikzpicture}
\caption{Uniform distributions.}
    \label{fig:illustration}
\vspace{-0mm}
\end{wrapfigure}
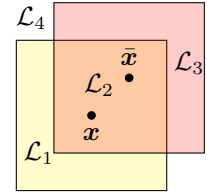

To illustrate the framework, we show a simple (but new) scenario when $\gX = \sR^d$ and $\phi$ is an additive uniform noise with a parameter $\gamma \in \sR_{>0}$: 
\begin{align}
\vspace{-1.5mm}
    \rand(\vx)_i = \vx_i + \veps_i, \veps_i \overset{i.i.d.}{\sim} \text{Uniform}([-\gamma, \gamma]), \forall i \in \{1,\dots,d\}. \label{eq:unif}
\end{align}
Given two points $\vx$ and $\bar \vx$, as illustrated in Fig.~\ref{fig:illustration}, we can partition the space $\sR^d$ into 4 disjoint regions: $\gL_1 = \gB_{\gamma, \infty}(\vx) \backslash \gB_{\gamma, \infty}(\bar \vx)$, $\gL_2 = \gB_{\gamma, \infty}(\vx) \cap \gB_{\gamma, \infty}(\bar \vx)$, $\gL_3 = \gB_{\gamma, \infty}(\bar \vx) \backslash \gB_{\gamma, \infty}(\vx)$ and $\gL_4 = \sR^d \backslash (\gB_{\gamma, \infty}(\bar \vx) \cup \gB_{\gamma, \infty}(\vx) )$. Accordingly, $\forall \bar f \in \gF$, we can rewrite $\Pr(\bar f(\phi (\vx)) = y)$ and $\Pr(\bar f(\phi (\bar \vx)) = y)$ as follows:
\begin{align}
    \Pr(\bar f(\phi (\vx)) = y) = \sum_{i=1}^4 \int_{\gL_i} \Pr(\phi(\vx) = \vz) \Pr(\bar f(\vz) = y)) \mathrm{d}\vz = \sum_{i=1}^4 \pi_i \int_{\gL_i} \Pr(\bar f(\vz) = y)) \mathrm{d}\vz, \nonumber\\
    \Pr(\bar f(\phi (\bar \vx)) = y) = \sum_{i=1}^4 \int_{\gL_i} \Pr(\phi(\bar \vx) = \vz) \Pr(\bar f(\vz) = y)) \mathrm{d}\vz = \sum_{i=1}^4 \bar \pi_i \int_{\gL_i} \Pr(\bar f(\vz) = y)) \mathrm{d}\vz, \nonumber
\end{align}
where $\pi_{1:4} = ((2\gamma)^{-d}, (2\gamma)^{-d}, 0, 0)$, and $\bar \pi_{1:4} = (0, (2\gamma)^{-d}, (2\gamma)^{-d}, 0)$. With this representation, it is clear that, in order to solve Eq.~(\ref{eqn:target}), we only have to consider the integral behavior of $\bar f$ within each region $\gL_1,\dots,\gL_4$. Concretely, we have:
\begin{align}
\vspace{-2.5mm}
    \rho_{\vx, \bar \vx}(p) & = \min_{\substack{\bar f \in \gF:\sum_{i=1}^4 \pi_i \int_{\gL_i} \Pr(\bar f(\vz) = y)) \mathrm{d}\vz = p }} \sum_{i=1}^4 \bar \pi_i \int_{\gL_i} \Pr(\bar f(\vz) = y)) \mathrm{d}\vz\nonumber \\
    & = \min_{\substack{g:\{1,2,3,4\} \to [0, 1],\\ \pi_1 |\gL_1|g(1) + \pi_2 |\gL_2| g(2) = p}} \bar \pi_2 |\gL_2| g(2) + \bar \pi_3 |\gL_3| g(3) = \min_{\substack{g:\{1,2,3,4\} \to [0, 1],\\ \pi_1 |\gL_1|g(1) + \pi_2 |\gL_2| g(2) = p}} \bar \pi_2 |\gL_2| g(2), \nonumber
\end{align}
where the second equality filters the components with $\pi_i = 0$ or $\bar \pi_i = 0$, and the last equality is due to the fact that $g(3)$ is unconstrained and minimizes the objective when $g(3) = 0$. Since $\pi_2 = \bar \pi_2$, 
\begin{align*}
\vspace{-1.5mm}
    \begin{cases}
    \rho_{\vx, \bar \vx}(p)  = 0, &\text{ if } 0 \leq p \leq \pi_1 |\gL_1| = \Pr(\phi(\vx) \in \gL_1),\\
    \rho_{\vx, \bar \vx}(p)  = p - \pi_1 |\gL_1|, &\text{ if } 1 \geq p > \pi_1 |\gL_1| = \Pr(\phi(\vx) \in \gL_1).
    \end{cases}
\vspace{-0.5mm}
\end{align*}
To obtain the regional certificate, the minimizers of $\min_{\bar \vx \in \gB_{r, q}(\vx)} \rho_{\vx, \bar \vx}(p)$ are simply the points that maximize the volume of $\gL_1 = \gB_1 \backslash \gB_2$. Accordingly,
\begin{proposition}
If $\rand(\cdot)$ is defined as Eq.~(\ref{eq:unif}), we have $R(\vx, p, q=1) = {2p \gamma - } \gamma$ and $R(\vx, p, q=\infty) = {2\gamma - } 2\gamma(1.5 - p)^{1/d}$.
\label{prop:uniform}
\end{proposition}

\header{Discussion.} \textblue{Our goal here was to illustrate how certificates can be computed with the uniform distribution using our technique. However, the certificate radius itself is inadequate in this case. For example, $R(\vx, p, q=1) \leq \gamma$, which arises from the bounded support in the uniform distribution.} The derivation nevertheless provides some insights about how one can compute the point-wise certificate $\rho_{\vx, \bar \vx} (p)$. The key step is to partition the space into regions $\gL_1,\dots,\gL_4$, where the likelihoods $\Pr(\phi(\vx) = \vz)$ and $\Pr(\phi(\bar \vx) = \vz)$ are both constant within each region $\gL_i$. The property allows us to substantially reduce the optimization problem in Eq.~(\ref{eqn:target}) to finding a single probability value $g(i) \in [0, 1]$ for each region $\gL_i$.

\subsection{A General Lemma for Point-wise Certificate}

In this section, we generalize the idea in \xref{sec:warmup} to find the point-wise certificate $\rho_{\vx, \bar \vx}(p)$. For each point $\vz \in \gX$, we define the likelihood ratio $\eta_{\vx, \bar \vx}(\vz)\triangleq \Pr(\rand(\vx) = \vz) / \Pr(\rand(\bar \vx) = \vz)$.\footnote{If $\Pr(\rand(\bar \vx) = \vz) = \Pr(\rand(\vx) = \vz) = 0$, $\eta_{\vx, \bar \vx}(\vz)$ can be defined arbitrarily in $[0, \infty]$ without affecting the solution in Lemma~\ref{thm:main}.}  If we can partition $\gX$ into $n$ regions $\gL_1, \dots, \gL_n: \cup_{i=1}^n \gL_i = \gX$ for some $n \in \sZ_{>0}$, such that the likelihood ratio within each region $\gL_i$ is a constant $\eta_i \in [0, \infty]$: $\eta_{\vx, \bar \vx}(\vz) = \eta_i, \forall \vz \in \gL_i$, then we can sort the regions such that $\eta_1 \geq \eta_2 \geq \dots \geq \eta_n$. Note that $\gX$ can still be uncountable (see the example in \xref{sec:warmup}). 


\textblue{Informally, we can always ``normalize'' $\bar f$ so that it predicts a constant probability value $g(i) \in [0, 1]$ within each likelihood ratio region $\gL_i$. This preserves the integral over $\gL_i$ and thus over $\gX$, generalizing the scenario in \xref{sec:warmup}. Moreover, to minimize $\Pr(\bar f(\phi(\bar \vx)) = y)$ under a fixed budget $\Pr(\bar f(\phi(\vx)) = y)$, as in Eq.~(\ref{eqn:target}), it is advantageous to 
set $\bar f(\vz)$ to $y$ in regions with high likelihood ratio. These arguments suggest a greedy algorithm for solving Eq.~(\ref{eqn:target}) by iteratively assigning $f(\vz) = y, \forall \vz \in \gL_i$ for $i \in (1,2,\dots)$ until the budget constraint is met.}
Formally,
\begin{lemma}
\label{thm:main}
$\forall \vx, \bar \vx \in \gX$, $p \in [0, 1]$, let
$H^* \triangleq \min_{H \in \{1,\dots,n\}: \sum_{i=1}^H \Pr(\rand(\vx) \in \gL_i) \geq p} H,$
then $\eta_{H^*}>0$, any $f^*$ satisfying Eq.~(\ref{eqn:optimal_f}) is a minimizer of Eq.~(\ref{eqn:target}),
\begin{align}
    \forall i \in \{1,2,\dots,n\}, \forall \vz \in \gL_i, & \Pr(f^*(\vz) = y) =\begin{cases}
         1,  &\text{if } i < H^*,\\
        \frac{p - \sum_{i=1}^{H^*-1} \Pr(\rand(\vx) \in \gL_i)}{\Pr(\rand(\vx) \in \gL_{H^*})},  &\text{if } i = H^*,\\
         0, &\text{if } i > H^*.
        \end{cases} \label{eqn:optimal_f}
\end{align}
and
$\rho_{\vx, \bar \vx}(p)
    = \sum_{i = 1}^{H^* - 1} \Pr(\rand(\bar \vx) \in \gL_i) + (p - \sum_{i=1}^{H^*-1} \Pr(\rand(\vx) \in \gL_i)) / \eta_{H^*}$
\end{lemma}

We remark that Eq.~(\ref{eqn:target}) and Lemma~\ref{thm:main} can be interpreted as a likelihood ratio testing~\cite{neyman1933ix}, by casting $\Pr(\phi(\vx) = \vz)$ and $\Pr(\phi(\bar \vx) = \vz)$ as likelihoods for two hypothesis with the significance level $p$. We refer the readers to \cite{tocher1950extension} to see a similar Lemma derived under the language of hypothesis testing. 


\begin{remark}
$\rho_{\vx, \bar \vx}(p)$  is an increasing continuous function of $p$; if $\eta_1 < \infty$, $\rho_{\vx, \bar \vx}(p)$ is a strictly increasing continuous function of $p$; if $\eta_1 < \infty$ and $\eta_n > 0$, 
$\rho_{\vx, \bar \vx}:[0,1]\to[0,1]$ is a bijection.

\label{rmk:new:bijection}
\end{remark}
Remark~\ref{rmk:new:bijection} will be used in \xref{sec:l0-perturb} to derive an efficient algorithm to compute robustness certificates. 

\header{Discussion.} Given $\gL_i$, $\Pr(\phi(\vx) \in \gL_i)$, and $\Pr(\phi(\bar \vx) \in \gL_i), \forall i \in [n]$, Lemma~\ref{thm:main} provides an $O(n)$ method to compute $\rho_{\vx, \bar \vx}(p)$. For any actual randomization $\rand$, the key is to find a partition $\gL_1,\dots,\gL_n$ such that $\Pr(\phi(\vx) \in \gL_i)$ and $\Pr(\phi(\bar \vx) \in \gL_i)$ are easy to compute. Having constant likelihoods in each $\gL_i: \Pr(\phi(\vx)=\vz) = \Pr(\phi(\vx)=\vz'), \forall \vz, \vz' \in \gL_i$ (cf. only having constant likelihood ratio $\eta_i$) is a way to simplify $\Pr(\phi(\vx) \in \gL_i) = |\gL_i| \Pr(\phi(\vx)=\vz)$, and similarly for $\Pr(\phi(\bar \vx) \in \gL_i)$.

\subsection{A Discrete Distribution for \texorpdfstring{$\ell_0$}{l0} Robustness}
\label{sec:l0-perturb}
We consider $\ell_0$ robustness guarantees in a discrete space $\gX = \left\{0, \frac1K, \frac2K,\dots, 1\right\}^d $ for some $K \in \sZ_{>0}$;\footnote{\textgrey{More generally, the method applies to the $\ell_0$ / Hamming distance in a Hamming space (i.e., fixed length sequences of tokens from a discrete set, e.g., $(\spadesuit 10, \spadesuit J, \spadesuit Q, \spadesuit K, \spadesuit A) \in \{\spadesuit A, \spadesuit K,...,\clubsuit 2\}^{5}$).}} we define the following discrete distribution with a parameter $\alpha \in (0,1)$, independent and identically distributed for each dimension $i \in \{1,2,\dots,d\}$:
\begin{align}
    \begin{cases}
    \Pr(\rand(\vx)_i = \vx_i) = \alpha, &\\
    \Pr(\rand(\vx)_i = z) = (1-\alpha) / K \triangleq \beta\in (0,1/K), &\text{if } z \in \left\{0, \frac1K,
\frac2K,\dots, 1\right\} \text{ and } z \neq \vx_i.
    \end{cases}
    \label{eqn:perturb}
\end{align}
Here $\rand(\cdot)$ can be regarded as a composition of a Bernoulli random variable and a uniform random variable.  Due to the symmetry of the randomization with respect to all the configurations of $\vx, \bar \vx \in \gX$ such that $\|\vx - \bar \vx\|_0 =r$ (for some $r \in \sZ_{\geq0}$), we have the following Lemma for the equivalence of $\rho_{\vx, \bar \vx}$: 
\begin{lemma}
\label{lem:canonical}
\textcyan{
If $\rand(\cdot)$ is defined as Eq.~(\ref{eqn:perturb}), given $r\in \sZ_{\geq 0}$,  define the canonical vectors $\vx_C\triangleq (0, 0, \cdots,  0)$ and $\bar \vx_C\triangleq (1, 1, \cdots, 1, 0, 0, \cdots, 0)$, where $\|\bar \vx_C \|_0=r$. Let $\rho_r\triangleq \rho_{\vx_C, \bar \vx_C}$.  Then for all $\vx, \bar \vx$ such that $ \|\vx- \bar \vx\|_0 =r$, we have $\rho_{\vx, \bar \vx}=\rho_r$.}

\end{lemma}

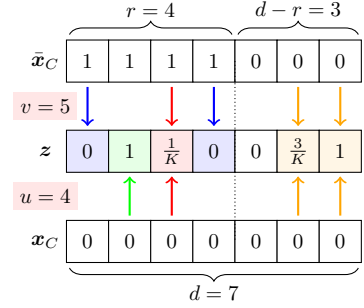
\begin{wrapfigure}{r}{0.335\textwidth}
  \begin{center}
 \vspace{-0.9cm}
	\begin{tikzpicture}[scale=0.8, transform shape]
	\foreach \x/\y/\n/\c/\f in 
	{0/0/1/black/0,1/0/1/black/0,2/0/1/black/0,3/0/1/black/0,4/0/0/black/0,5/0/0/black/0,6/0/0/black/0,
0/1/0/dodgerblue/10,1/1/1/orange/10,2/1/\frac1K/tomato/10,3/1/0/dodgerblue/10,4/1/0/black/0,5/1/\frac3K/lime/10,6/1/1/lime/10,
0/2/0/black/0,1/2/0/black/0,2/2/0/black/0,3/2/0/black/0,4/2/0/black/0,5/2/0/black/0,6/2/0/black/0}{
		\node[draw=black,fill=\c!\f,rectangle,minimum height = 0.7cm,
			minimum width=0.7cm] at (0.7*\x+0,1.5-1.5*\y) {$\n$};
	}
	\node at (-0.7,1.5) {$\bar\vx_C$};
	\node at (-0.7,0) {$\vz$};
	\node at (-0.7,-1.5) {$ \vx_C$};
	\node[fill=red!10] at (-0.7,-0.75) {$u=4$};
	\node[fill=red!10] at (-0.7,0.75) {$v=5$};
	
	\draw[->,thick, dodgerblue ] (0,1.07)--+(0,-0.65);
	\draw[->,thick, tomato ] (0.7*2,1.07)--+(0,-0.65);
	\draw[->,thick ,dodgerblue] (0.7*3,1.07)--+(0,-0.65);
	\draw[->,thick ,lime] (0.7*5,1.07)--+(0,-0.65);
	\draw[->,thick,lime ] (0.7*6,1.07)--+(0,-0.65);

\draw[densely dotted] (0.7*3+0.35, 1.5) --++(0, -3);
\draw [decorate,decoration={brace,amplitude=5pt,raise=4ex}]
(-0.3,1.15) --+ (2.7,0) node[midway,yshift=1.2cm]{$r=4$};
\draw [decorate,decoration={brace,amplitude=5pt,raise=4ex}]
(2.5,1.15) --+ (2,0) node[midway,yshift=1.2cm]{$d-r=3$};
\draw [decorate,decoration={brace,amplitude=5pt,mirror,raise=4ex}]
(-0.3,-1.15) --+ (4.8,0) node[midway,yshift=-1.2cm]{$d=7$};

	\draw[->,thick,orange ] (0.7*1,-1.07)--+(0,0.65);
\draw[->,thick, tomato ] (0.7*2,-1.07)--+(0,0.65);
\draw[->,thick ,lime] (0.7*5,-1.07)--+(0,0.65);
\draw[->,thick,lime ] (0.7*6,-1.07)--+(0,0.65);
	\end{tikzpicture}
	 \vspace{-0.55cm}
\end{center}
  \caption{Illustration for Eq.~(\ref{eqn:LMN})}
  \label{fig:flip_illustration}
  	 \vspace{-0.5cm}
\end{wrapfigure}

Based on Lemma~\ref{lem:canonical}, finding $R(\vx, p, q)$ for a given $p$, it suffices to find the maximum $r$ such that $\rho_r(p)>0.5$.  Since the likelihood ratio $\eta_{\vx, \bar \vx}(\vz)$ is always positive and finite, the inverse $\rho_r^{-1}$ exists (due to Remark~\ref{rmk:new:bijection}), which allows us to pre-compute $\rho_r^{-1}(0.5)$ and check $p > \rho_r^{-1}(0.5)$ for each $r \in \sZ_{\geq 0}$, instead of computing $\rho_r(p)$ for each given $p$ and $r$.  Then $R(\vx, p, q)$ is simply the maximum $r$ such that $p > \rho_r^{-1}(0.5)$.  Below we discuss how to compute $\rho^{-1}_r(0.5)$ in a scalable way.  Our first step is to identify a set of likelihood ratio regions $\gL_1,\dots,\gL_n$ such that $\Pr(\phi(\vx)\in \gL_i)$ and $\Pr(\phi(\bar \vx)\in \gL_i)$ as used in Lemma~\ref{thm:main} can be computed efficiently. Note that, due to Lemma~\ref{lem:canonical}, \textcyan{it suffices to consider $\vx_C, \bar \vx_C$ such that $\|\bar \vx_C\|_0=r$ throughout the derivation. }

For an $\ell_0$ radius $r \in \sZ_{\geq 0}$, $\forall (u, v) \in \{0, 1, \dots, d\}^2$, we construct the region
\begin{equation}
\label{eqn:LMN}
\gL(u, v; r) \triangleq \{\vz \in \gX: \Pr(\phi(\vx_C) = \vz) = \alpha^{d-u} \beta^u, \Pr(\phi(\bar \vx_C) = \vz) = \alpha^{d-v} \beta^v\},    
\end{equation}
which \textcyan{contains points that can be obtained by ``flipping'' $u$ coordinates from $\vx_C$ or $v$ coordinates from $\bar \vx_C$. See Figure \ref{fig:flip_illustration} for an illustration, where different colors represent different types of coordinates: orange means both $\vx_C, \bar \vx_C$ are flipped on this coordinate and they were initially the same; red means both are flipped and were initially different; green means only $ \vx_C$ is flipped and blue means only $\bar \vx_C$ is flipped. By denoting the numbers of these coordinates as $i,j^*,u-i-j^*, v-i-j^*$, respectively, we have the following formula for computing the cardinality of each region $|\gL(u, v; r)|$.}
\begin{lemma}
\label{lem:computation}
For any $u, v \in \{0, 1, \dots, d\}, u \leq v, r \in \sZ_{\geq 0}$ we have  $|\gL(u, v; r)| = |\gL(v, u; r)|$, and 
\begin{align*}
    |\gL(u, v; r)| & = \sum_{i=\max\{0, v - r\}}^{\min\left(u, d-r, \lfloor \frac{u+v-r}{2} \rfloor \right)} \frac{(K - 1)^{j^*} r !}{(u-i-j^*)!(v-i-j^*)!j^*!} \frac{K^i(d - r)!}{(d-r-i)! i!},  \\
    & \text{where } j^* \triangleq u + v - 2i - r.
\end{align*}
\end{lemma}
\begin{wrapfigure}{R}{0.36\textwidth}
\vspace{-4mm}
\begin{minipage}{0.36\textwidth}
\begin{algorithm}[H]
\begin{algorithmic}[1]
\STATE 
 sort $\{(u_i, v_i)\}_{i=1}^{n}$ by likelihood ratio
\STATE $p, \rho_r = 0, 0$\;
 \FOR{$i = 1,\dots,n$}
  \STATE $p'    = \alpha^{d - u_i} \beta^{u_i}$\;
  \STATE $\rho_r' = \alpha^{d - v_i} \beta^{v_i}$\;
  \STATE $\Delta \rho_r = \rho_r' \times |\gL(u_i, v_i; r)|$\;
  \IF {$\rho_r + \Delta \rho_r < 0.5$}
 \STATE  $\rho_r =\rho_r+ \Delta \rho_r$\;
 \STATE  $p =p+ p' \times |\gL(u_i, v_i; r)|$\;
 \ELSE
 \STATE  $p =p+ p' \times (0.5 - \rho_r) / \rho_r'$\!\!\!\!\!\!\;
 \STATE  return $p$
 \ENDIF
 \ENDFOR
 \end{algorithmic}
\caption{Computing $\rho_r^{-1}(0.5)$}
\label{alg}
\end{algorithm}
\end{minipage}
\vspace{-5mm}
\end{wrapfigure}
\textcyan{Therefore, for a fixed $r$, the complexity of computing all the cardinalities $|\gL(u, v; r)|$ is $\Theta(d^3)$.}  Since each region $\gL(u, v; r)$ has a constant likelihood ratio $\alpha^{v-u}\beta^{u-v}$ and we have $\cup_{u=0}^d \cup_{v=0}^d \gL(u, v; r) = \gX$, we can apply the regions to find the function $\rho_{\vx, \bar \vx} = \rho_{r}$ via Lemma~\ref{thm:main}.  Under this representation, the number of nonempty likelihood ratio regions $n$ is bounded by $(d+1)^2$, the perturbation probability $\Pr(\phi(\vx) \in \gL(u, v; r))$ used in Lemma~\ref{thm:main} is simply $\alpha^{d-u} \beta^u|\gL(u, v; r)|$, and similarly for the $\Pr(\phi(\bar \vx) \in \gL(u, v; r))$.  Based on Lemma~\ref{thm:main} and Lemma~\ref{lem:computation}, we may use a for-loop to compute the bijection $\rho_r(\cdot)$ for the input $p$ until $\rho_r(p) = 0.5$, and return the corresponding $p$ as $\rho_r^{-1}(0.5)$. The procedure is illustrated in Algorithm~\ref{alg}.

\header{Scalable implementation.}
In practice, Algorithm \ref{alg} can be challenging to implement; the probability values (e.g., $\alpha^{d-u} \beta^{u}$) can be extremely small, which is infeasible to be computationally represented using floating points.  If we set $\alpha$ to be a rational number, both $\alpha$ and $\beta$ can be represented in fractions, and thus all the corresponding probability values can be represented by two (large) integers; we also observe that computing the (large) cardinality $|\gL(u, v; r)|$ is feasible in modern large integer computation frameworks in practice (e.g., \texttt{python}), which motivates us to adapt the computation in Algorithm~\ref{alg} to large integers.

For simplicity, we assume $\alpha = \alpha' / 100$ with some $\alpha' \in \sZ: 100 \geq \alpha' \geq 0$.  If we define $\tilde \alpha \triangleq 100K \alpha \in \sZ, \tilde \beta \triangleq 100K \beta \in \sZ$, we may implement Algorithm~\ref{alg} in terms of the non-normalized, integer version $\tilde \alpha, \tilde \beta$.  Specifically, we replace $\alpha, \beta$ and the constant $0.5$ with $\tilde \alpha, \tilde \beta$ and $50K\times(100K)^{d-1}$, respectively.  Then all the computations in Algorithm~\ref{alg} can be trivially adapted except the division $(0.5 - \rho_r) / \rho'_r$.  Since the division is bounded by $|\gL(u_i, v_i; r)|$ (see the comparison between line 9 and line 11), we can implement the division by a binary search over $\{1,2\dots, |\gL\{m_i, n_i\}|\}$, which will result in an upper bound with an error bounded by $\rho'_r$ in the original space, which is in turn bounded by $\alpha^d$ assuming $\alpha > \beta$.  Finally, to map the computed, unnormalized $\rho^{-1}_r (0.5)$, denoted as $\tilde \rho^{-1}_r (0.5)$, back to the original space, we find an upper bound of $\rho^{-1}_r (0.5)$ up to the precision of $10^{-c}$ for some $c \in \sZ_{>0}$ (we set $c=20$ in the experiments): we find the smallest upper bound of $\tilde \rho^{-1}_r (0.5) \leq \hat \rho \times (10K)^c (100K)^{d-c}$ over $\hat \rho \in \{1, 2, \dots, 10^c\}$ via binary search, and report an upper bound of $\rho_r^{-1}(0.5)$ as $\hat \rho \times 10^{-c}$ with an error bounded by $10^{-c} + \alpha^d$ in total. Note that an upper bound of $\rho_r^{-1}(0.5)$ is still a valid certificate.

As a side note, simply computing the probabilities in the log-domain will lead to uncontrollable approximate results due to floating point arithmetic; using large integers to ensure a verifiable approximation error in Algorithm \ref{alg} is necessary to ensure a computationally accurate certificate.

\subsection{Connection Between the Discrete Distribution and an Isotropic Gaussian Distribution}
\label{sec:comparison}

When the inputs are binary vectors $\gX = \{0, 1\}^d$, one may still apply the prior work~\cite{cohen2019certified} using an additive isotropic Gaussian noise $\phi$ to obtain an $\ell_0$ certificates since there is a bijection between $\ell_0$ and $\ell_2$ distance in $\{0, 1\}^d$. If one uses a denoising function $\zeta(\cdot)$ that projects each randomized coordinate $\rand(\vx)_i \in \sR$ back to the space $\{0, 1\}$ using the (likelihood ratio testing) rule
\begin{align}
    \zeta(\rand(\vx))_i = \one\{\rand(\vx)_i > 0.5\}, \forall i \in [d], 
    \nonumber
\end{align}
then the composition $\zeta\circ \rand$ is equivalent to our discrete randomization scheme with $\alpha = \Phi(0.5; \mu=0, \sigma^2),$ where $\Phi$ is the CDF function of the Gaussian distribution with mean $\mu$ and variance $\sigma^2$.

If one applies a classifier upon the composition (or, equivalently, the discrete randomization scheme), then the certificates obtained via the discrete distribution is always tighter than the one via Gaussian distribution. Concretely, we denote $\gF_{\zeta} \subset \gF$ as the set of measurable functions with respect to the Gaussian distribution that can be written as the composition $\bar f' \circ \zeta$ for some $\bar f'$, and we have
\begin{align}
    \min_{\bar f \in \gF_{\zeta}: \Pr(\bar f(\rand(\vx)) = y) = p}
    \Pr(\bar f(\rand(\bar{\vx})) = y) \geq 
    \min_{\bar f \in \gF: \Pr(\bar f(\rand(\vx)) = y) = p}
    \Pr(\bar f(\rand(\bar{\vx})) = y),
    \nonumber
\end{align}
where the LHS corresponds to the certificate derived from the discrete distribution (i.e., applying $\zeta$ to an isotropic Gaussian), and the RHS corresponds to the certificate from the Gaussian distribution. 

\subsection{A Certificate with Additional Assumptions}
\label{sec:refinement}

In the previous analyses, we assume nothing but the measurability of the classifier. If we further make assumptions about the functional class of the classifier, we can obtain a tighter certificate than the ones outlined in \xref{sec:framework}. Assuming an extra denoising step in the classifier over an additive Gaussian noise as illustrated in \xref{sec:comparison}  is one example.

Here we illustrate the idea with another example. We assume that the inputs are binary vectors $\gX = \{0, 1\}^d$, the outputs are binary $\gY = \{0, 1\}$, and that the classifier is a decision tree that each input coordinate can be used at most once in the entire tree. Under the discrete randomization scheme, the prediction probability can be computed via tree recursion, since a decision tree over the discrete randomization scheme can be interpreted as assigning a probability of visiting the left child and the right child for each decision node.  To elaborate, we denote $\texttt{idx}[i], \texttt{left}[i],$ and $\texttt{right}[i]$ as the split feature index, the left child and the right child of the $i^\text{th}$ node.  Without loss of generality, we assume that each decision node $i$ routes its input to the right branch if $\vx_{\texttt{idx}[i]} = 1$.  Then $\Pr(f(\phi(\vx))=1)$ can be found by the recursion
\begin{align}
    \texttt{pred}[i] = \alpha^{\one\{\vx_{\texttt{idx}[i]} = 1\}}\beta^{\one\{\vx_{\texttt{idx}[i]} = 0\}} \texttt{pred}[\texttt{right}[i]] + \alpha^{\one\{\vx_{\texttt{idx}[i]} = 0\}}\beta^{\one\{\vx_{\texttt{idx}[i]} = 1\}} \texttt{pred}[\texttt{left}[i]], \label{eq:dp}
\end{align}
where the boundary condition is the output of the leaf nodes. \textred{Effectively}, we are recursively aggregating the partial solutions found in the left subtree and the right subtree rooted at each node $i$, and $\texttt{pred}[\texttt{root}]$ is the final prediction probability.   Note that changing one input coordinate in $\vx_k$ is equivalent to changing the recursion in the corresponding unique node $i'$ (if exists) that uses feature $k$ as the splitting index, which gives
\begin{align}
    \texttt{pred}[i'] = \alpha^{\one\{\vx_{\texttt{idx}[i']} = 0\}}\beta^{\one\{\vx_{\texttt{idx}[i']} = 1\}} \texttt{pred}[\texttt{right}[i']] + \alpha^{\one\{\vx_{\texttt{idx}[i']} = 1\}}\beta^{\one\{\vx_{\texttt{idx}[i']} = 0\}} \texttt{pred}[\texttt{left}[i']]. 
    \nonumber
\end{align}
\textred{
In addition, changes in the left subtree do not affect the partial solution found in the right subtree, and vice versa.} Hence, we may use dynamic programming to find the \emph{exact} adversary under each $\ell_0$ radius $r$ by aggregating the worst case changes found in the left subtree and the right subtree rooted at each node~$i$. \textcyan{See Appendix~\ref{sec:dp_tree} for details.}

\section{Learning and Prediction in Practice}

Since we focus on the development of certificates, here we only briefly discuss how we train the classifiers and compute the prediction probability $\Pr(f(\phi(\vx))=y)$ in practice.

\header{Deep networks}: We follow the approach proposed by the prior work~\cite{lecuyer2018certified}: training is conducted on samples drawn from the randomization scheme via a cross entropy loss.  The prediction probability $\Pr(f(\phi(\vx))=y)$ is estimated by the lower bound of the Clopper-Pearson Bernoulli confidence interval~\cite{clopper1934use} with \textred{100K samples drawn from the distribution and the $99.9\%$ confidence level.} Since $\rho_{\vx, \bar \vx}(p)$ is an increasing function of $p$ (Remark~\ref{rmk:new:bijection}), a lower bound of $p$ entails a valid certificate.  

\header{Decision trees}: we train the decision tree greedily in a breadth-first ordering with a depth limit; for each split, we only search coordinates that are not used before to enforce the functional constraint in \xref{sec:refinement}, and optimize a weighted gini index, which weights each training example $\vx$ by the probability that it is routed to the node by the discrete randomization. \textcyan{The details of the training algorithm is in Appendix \ref{sec:tree_training}.} The prediction probability is computed by Eq.~(\ref{eq:dp}).

\section{Experiment}
\label{sec:exp}

In this section, we validate the robustness certificates of the proposed discrete distribution ($\gD$) in $\ell_0$ norm.  We compare to the state-of-the-art additive isotropic Gaussian noise ($\gN$)~\cite{cohen2019certified}, since an $\ell_0$ certificate with radius $r$ in $\gX = \{0, \frac1K, \dots, 1\}^d$ can be obtained from an $\ell_2$ certificate with radius $\sqrt{r}$. Note that the derived $\ell_0$ certificate from Gaussian distribution is still tight with respect to all the measurable classifiers (see Theorem 1 in \cite{cohen2019certified}).  We consider the following evaluation measures:
\vspace{-2mm}
\begin{itemize}[leftmargin=4mm]
    \item $\mu(R)$: the average certified $\ell_0$ radius $R(\vx, p, q)$ (with respect to the labels) across the testing set.
    \item ACC@$r$: the certified accuracy within a radius $r$ (the average $\one\{R(\vx, p, q) \geq r\}$ in the testing set).\!
\end{itemize}

\subsection{Binarized MNIST}
\label{sec:exp:mnist}

\begin{table*}
  \caption{Randomly smoothed CNN models on the MNIST dataset. The first two rows refer to the same model with certificates computed via different methods (see details in \xref{sec:comparison}).}\label{tab:mnist}
  \centering
  \begin{tabular}{lccccccccccccccccccc}
    \toprule
    \multirow{2}{*}{\small $\phi$} & \multirow{2}{*}{\small Certificate} & \multirow{2}{*}{\small $\mu(R)$} &
    \multicolumn{7}{c}{\small ACC@$r$}\\
    \cmidrule(lr{0pt}){4-10} 
    & & & \small $r=1$ & \small $r=2$ & \small $r=3$ & \small $r=4$ & \small $r=5$ & \small $r=6$ & \small $r=7$\\
    \midrule
    \small $\gD$ & \small $\gD$ & \bf \small 3.456 & \bf \small 0.921 	& \bf \small 0.774 	& \bf \small 0.539  & \bf \small 0.524 	& \bf \small 0.357	& \bf \small 0.202 		& \bf \small 0.097\\
    \small $\gD$ & \small $\gN$~\cite{cohen2019certified} & \small 1.799 & \small 0.830 	& \small 0.557 	& \small 0.272  & \small 0.119 	& \small 0.021	& \small 0.000 		& \small 0.000\\ 
    \small $\gN$      & \small $\gN$~\cite{cohen2019certified} & \small 2.378 & \small 0.884 	& \small 0.701 	& \small 0.464  & \small 0.252 	& \small 0.078	& \small 0.000 		& \small 0.000\\ 
    \bottomrule
  \end{tabular}
\end{table*}

We use a $55,000$/$5,000$/$10,000$ split of the MNIST dataset for training/validation/testing.  For each data point $\vx$ in the dataset, we binarize each coordinate by setting the threshold as $0.5$.  Experiments are conducted on randomly smoothed CNN models and \textcyan{the implementation details are in Appendix~\ref{sec:CNN_model}.}  

The results are shown in Table~\ref{tab:mnist}. For the same randomly smoothed CNN model (the $1^\text{st}$ and $2^\text{nd}$ rows in Table~\ref{tab:mnist}), our certificates are consistently {better} than the ones derived from the Gaussian distribution \textred{(see \xref{sec:comparison})}. The gap between the average certified radius is about $1.7$ in $\ell_0$ distance, and the gap between the certified accuracy can be as large as $0.4$.  Compared to the models trained with Gaussian noise (the $3^\text{rd}$ row in Table~\ref{tab:mnist}), our model is also consistently better in terms of the measures.

Since the above comparison between our certificates and the Gaussian-based certificates is \emph{relative}, we conduct an exhaustive search over all the possible adversary within $\ell_0$ radii $1$ and $2$ to study the tightness against the \emph{exact} certificate. The resulting certified accuracies at radii $1$ and $2$ are $0.954$ and $0.926$, respectively, which suggest that our certificate is reasonably tight when $r=1$ ($0.954$ vs. $0.921$), but still too pessimistic when $r=2$ ($0.926$ vs. $0.774$). The phenomenon is expected since {the certificate is based on} \emph{all the measurable functions} for the discrete distribution. A tighter certificate requires additional assumptions on the classifier such as the example in \xref{sec:refinement}.
\subsection{ImageNet}

\begin{table*}[t]
  \caption{The guaranteed accuracy of randomly smoothed ResNet50 models on ImageNet. }\label{tab:imagenet}
  \centering
  \begin{tabular}{cccccccccccccccccccc}
    \toprule
    \multirow{2}{*}{ $\phi$ and certificate} & \multicolumn{7}{c}{\small ACC@$r$}\\
    \cmidrule(lr{0pt}){2-8} 
    & \small $r=1$ & \small $r=2$ & \small $r=3$ & \small $r=4$ & \small $r=5$ & \small $r=6$ & \small $r=7$\\
    \midrule
    \small $\gD$ & \bf \small 0.538 & \bf \small 0.394 & \bf \small 0.338 & \bf \small 0.274 & \bf \small 0.234 & \bf \small 0.190	& \bf \small 0.176\\
    \small $\gN$~\cite{cohen2019certified} & \small 0.372 & \small 0.292 & \small 0.226 & \small 0.194 & \small 0.170 & \small 0.154	& \small 0.138\\ 
    \bottomrule
  \end{tabular}
\end{table*}

We conduct experiments on ImageNet~\cite{deng2009imagenet}, a large scale image dataset with $1,000$ labels.  Following common practice, we consider the input space $\gX = \{0, 1/255, \dots, 1\}^{224\times224\times 3}$ by scaling the images. We consider the same ResNet50 classifier~\cite{he2016deep} and learning procedure as Cohen et al.~\cite{cohen2019certified} with the only modification on the noise distribution.  \textcyan{The details and visualizations can be found in Appendix~\ref{sec:supp_imagenet}.} For comparison, we report the best guaranteed accuracy of each method for each $\ell_0$ radius $r$ in Table~\ref{tab:imagenet}. Our model outperforms the competitor by a large margin at $r=1$ ($0.538$ vs. $0.372$), and consistently outperforms the baseline across different radii. 

\begin{figure}[t]
\vspace{-1mm}
\centering
    \begin{subfigure}{.32\textwidth}
		\centering
		\includegraphics[width=1\linewidth]{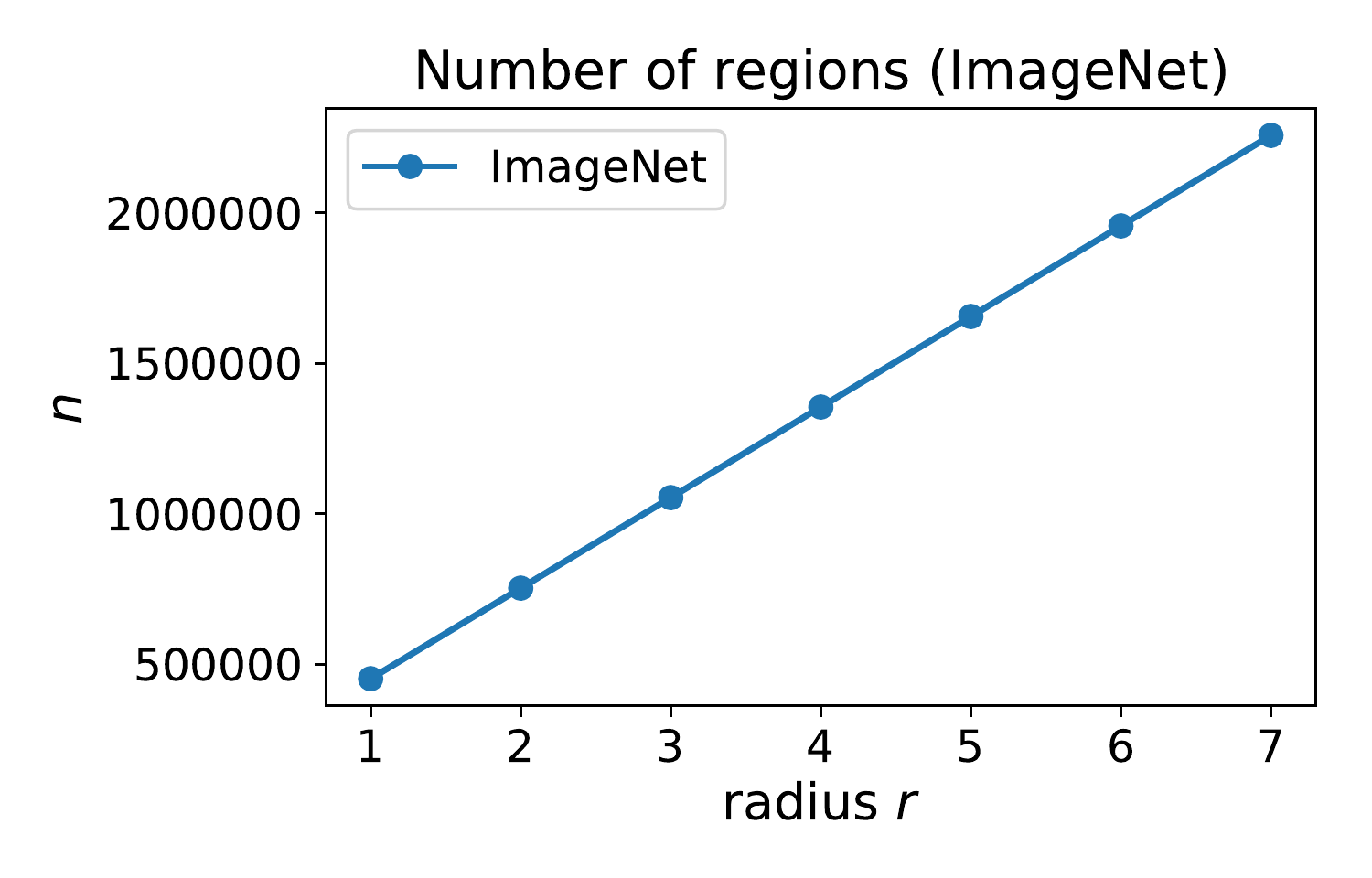}
		\caption{\small \# of nonempty $\gL(u, v; r)$}\label{fig:region}
	\end{subfigure}
	~
	\captionsetup[subfigure]{aboveskip=-0pt,belowskip=-0pt}
    \begin{subfigure}{.32\textwidth}
		\centering
		\includegraphics[width=1\linewidth]{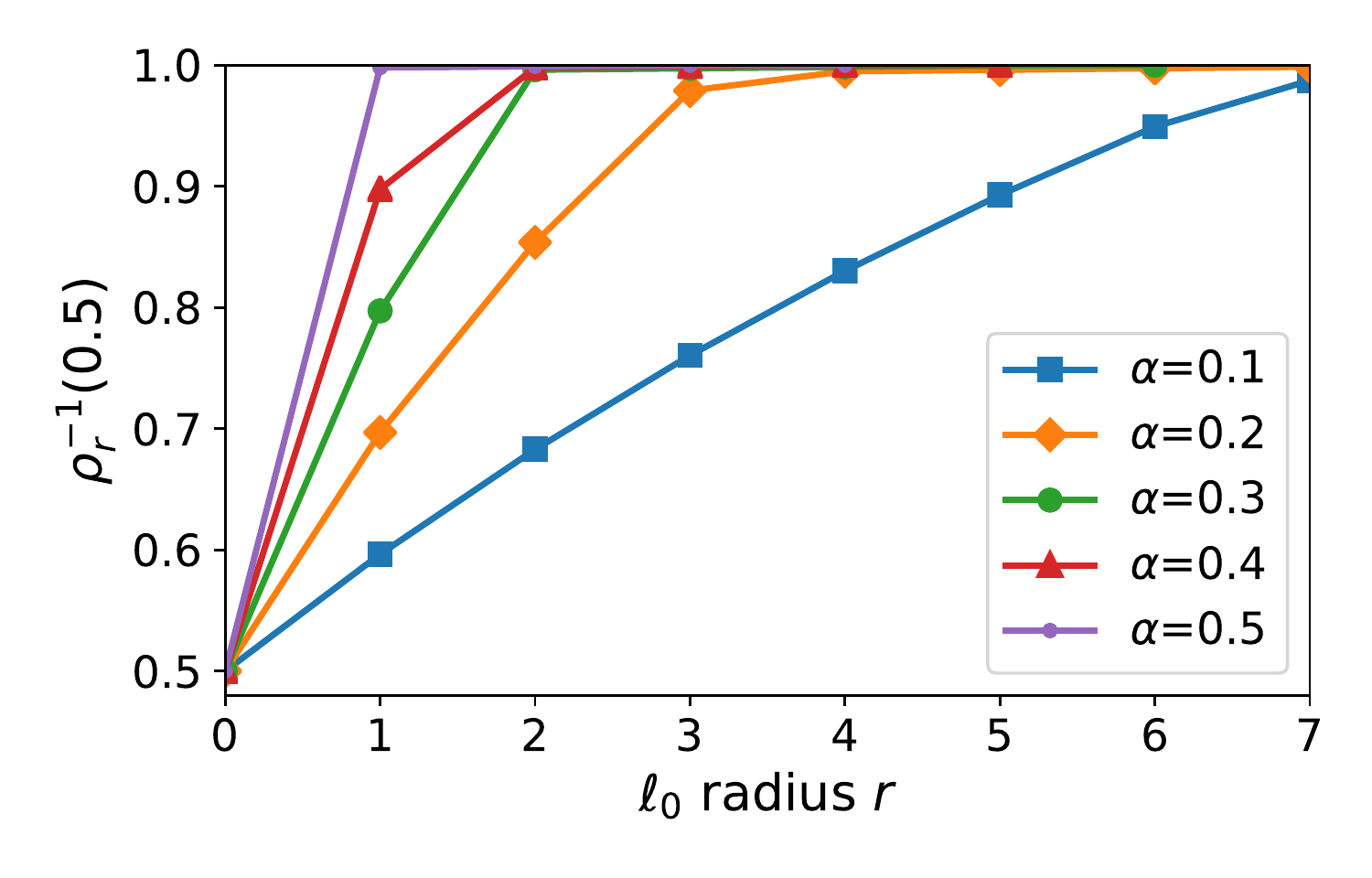}
		\caption{\small $\rho_r^{-1}(0.5)$ for an $\alpha$}\label{fig:threshold}
	\end{subfigure}
    ~
    \begin{subfigure}{.32\textwidth}
		\centering
		\includegraphics[width=1\linewidth]{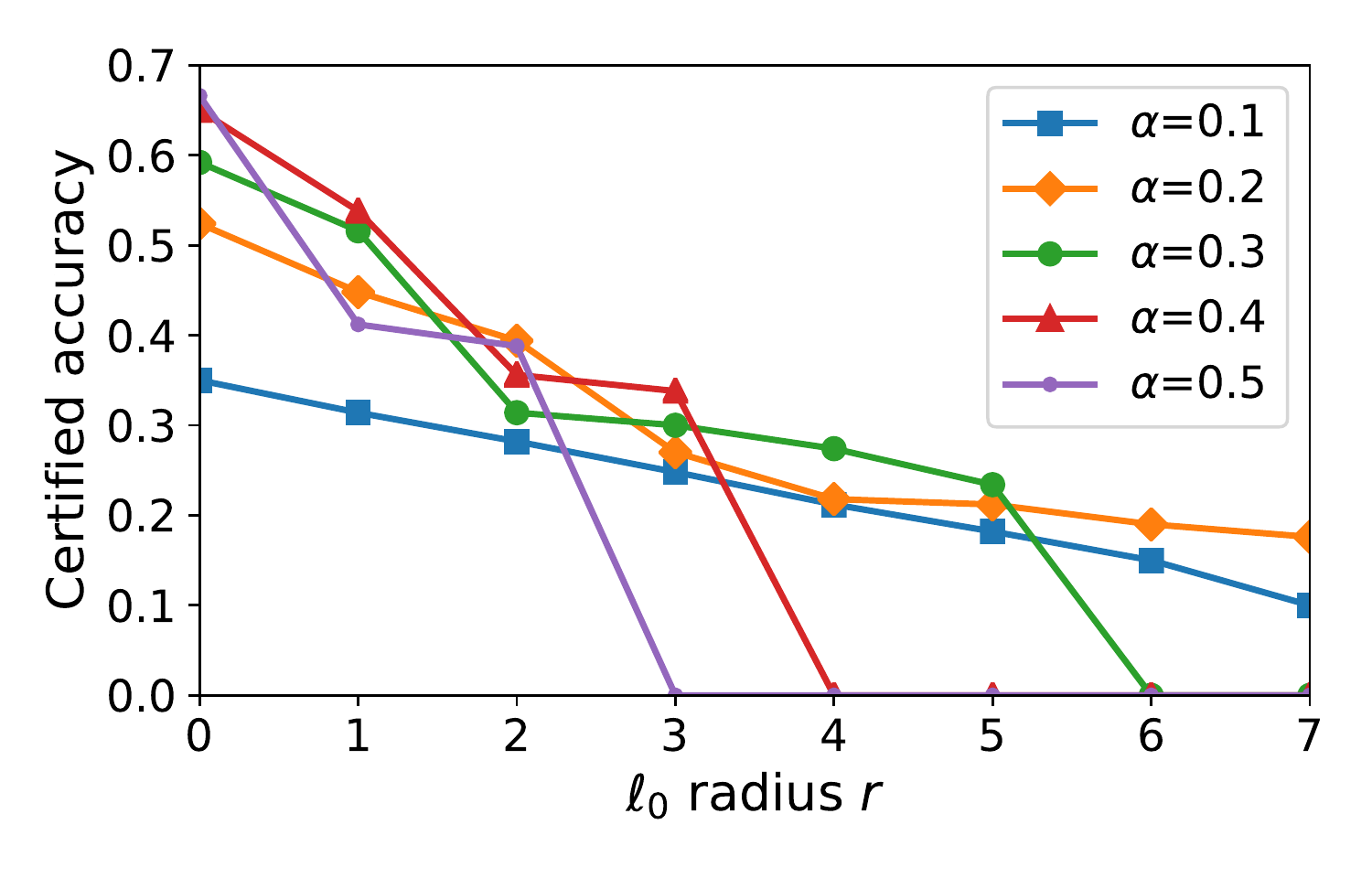}
		\caption{\small The certified accuracy for an $\alpha$}
	\end{subfigure}

    \caption{Analysis of the proposed method in the ImageNet dataset.}\label{fig:analysis}
    \vspace{-2.5mm}
\end{figure}

\begin{figure}[t]
\vspace{-1mm}
\centering
	\captionsetup[subfigure]{aboveskip=-0pt,belowskip=-0pt}
    \begin{subfigure}{.32\textwidth}
		\centering
		\includegraphics[width=1\linewidth]{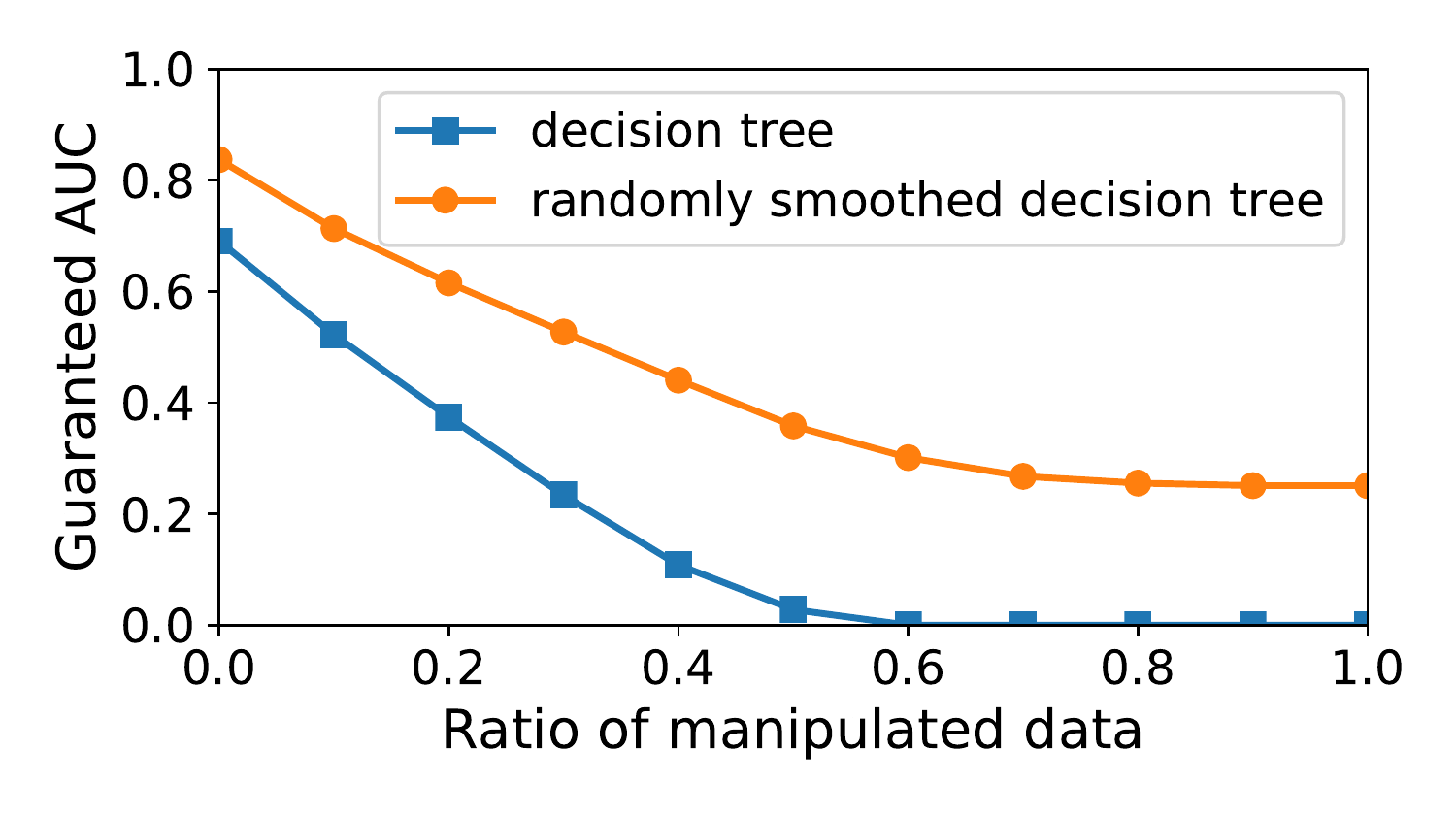}
		\caption{\small $r=1$}
	\end{subfigure}
    ~
    \begin{subfigure}{.32\textwidth}
		\centering
		\includegraphics[width=1\linewidth]{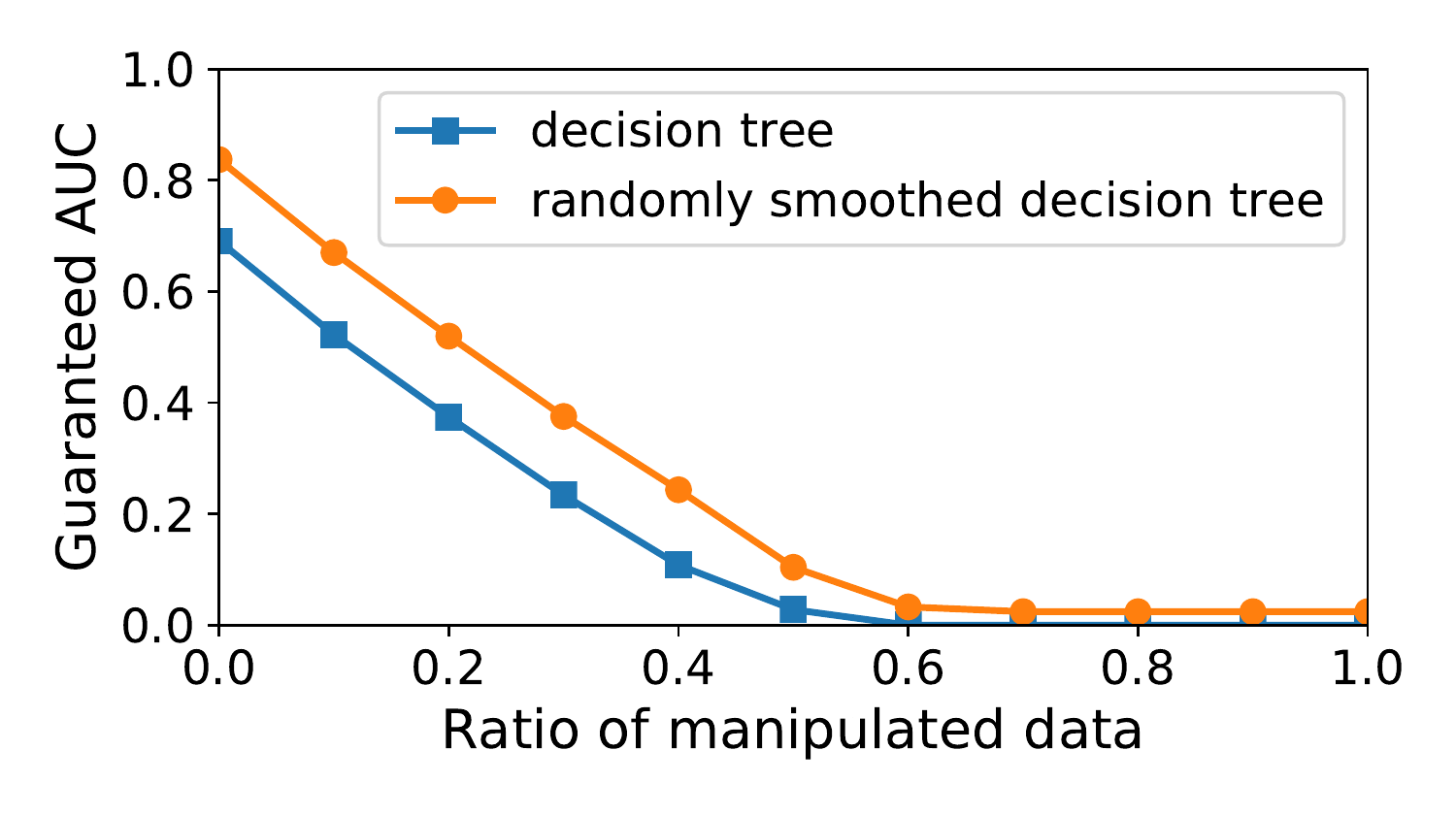}
		\caption{\small $r=2$}
	\end{subfigure}
	~
    \begin{subfigure}{.32\textwidth}
		\centering
		\includegraphics[width=1\linewidth]{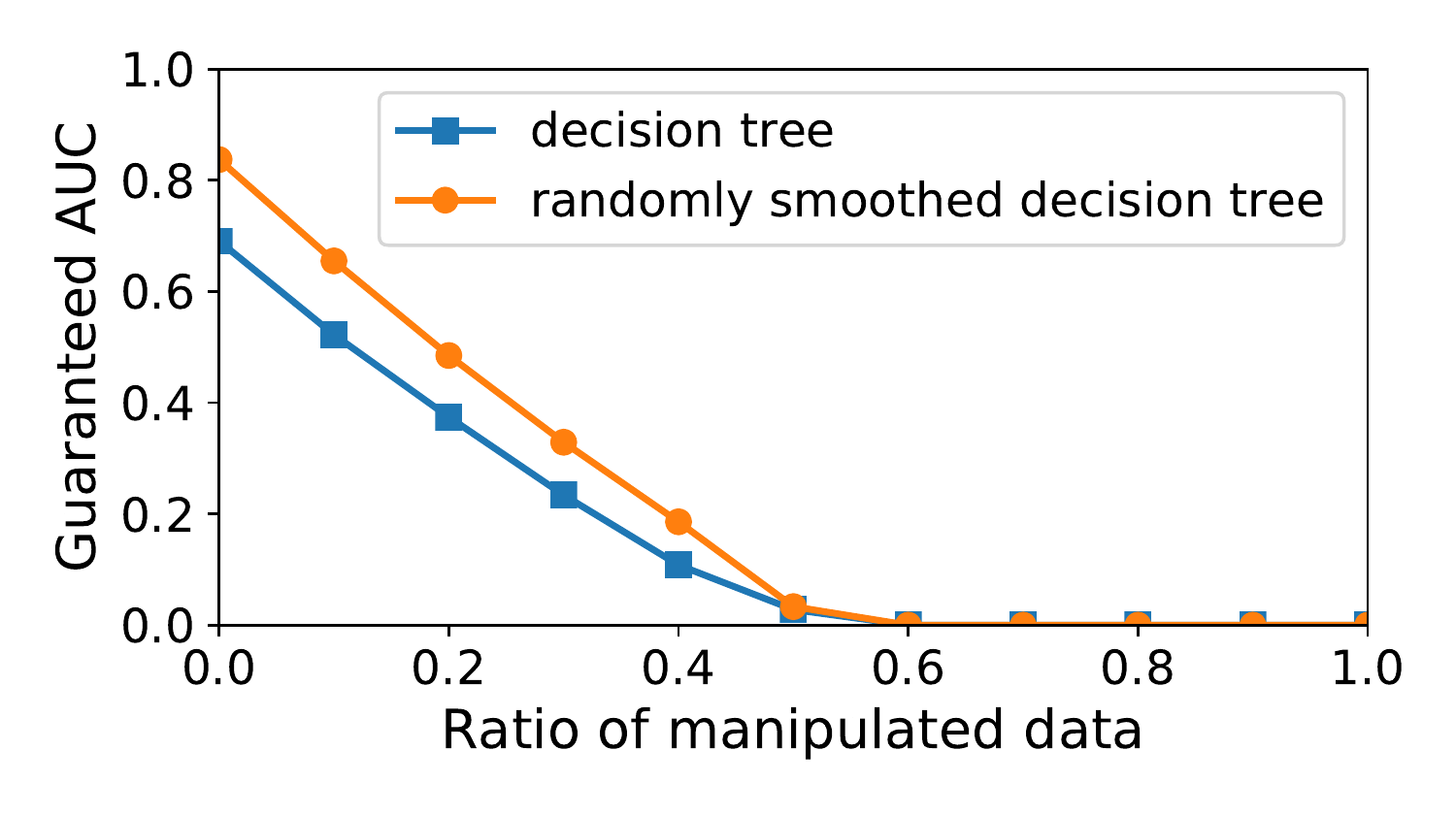}
		\caption{\small $r=3$}
	\end{subfigure}

    \caption{The guaranteed AUC in the Bace dataset across different $\ell_0$ radius $r$ and the ratio of testing data that the adversary can manipulate.}\label{fig:bace}
    \vspace{-2.5mm}
\end{figure}

\header{Analysis.} We analyze our method in ImageNet in terms of  1) the number $n$ of nonempty likelihood ratio region $\gL(u, v; r)$ in Algorithm~\ref{alg}, 2) the pre-computed $\rho_r^{-1}(0.5)$,  and 3) the certified accuracy at each $\alpha$. The results are in Figure~\ref{fig:analysis}. For reproducability, the detailed accuracy numbers of 3) is available in Table~\ref{tab:detail:imagenet} in Appendix~\ref{sec:supp_imagenet}, and the pre-computed $\rho_r^{-1}(0.5)$ is available at our code repository.
1) The number $n$ of nonempty likelihood ratio regions is much smaller than the bound $(d+1)^2 = (3 \times 224 \times 224)^2$ for small radii.  2) The value $\rho_r^{-1}(0.5)$ approaches $1$ more rapidly for a higher $\alpha$ value than a lower one. Note that $\rho_r^{-1}(0.5)$ only reaches $1$ when $r = d$ due to Remark~\ref{rmk:new:bijection}. Computing $\rho_r^{-1}(0.5)$ in large integer is time-consuming, which takes about $4$ days for each $\alpha$ and $r$, but this can be trivially parallelized across different $\alpha$ and $r$.\footnote{As a side note, computing $\rho_r^{-1}(0.5)$ in MNIST takes less than $1$ second for each $\alpha$ and $r$.} \textblue{For each radius $r$ and randomization parameter $\alpha$, note that the 4-day computation only has to be done \emph{once}, and the pre-computed $\rho^{-1}_r(0.5)$ can be applied to any ImageNet scale images and models.} 3) The certified accuracy behaves nonlinearly across different radii; relatively, a high $\alpha$ value exhibits a high certified accuracy at small radii and low certified accuracy at large radii, and vice versa.

\subsection{Chemical Property Prediction}

The experiment is conducted on the Bace dataset~\cite{subramanian2016computational}, a binary classification dataset for biophysical property prediction \textred{on molecules.} We use the Morgan fingerprints~\cite{rogers2010extended} to represent molecules, which are commonly used binary features~\cite{wu2018moleculenet} indicating the presence of various chemical substructures. The dimension of the features (fingerprints) is $1,024$. Here we focus on an ablation study comparing the proposed randomly smoothed decision tree with a vanilla decision tree, where the adversary is found by dynamic programming in \xref{sec:refinement} (thus the exact worse case) and a \textcyan{greedy} search, respectively. \textcyan{More details can be found in Appendix~\ref{sec:supp_chemical}}. 

Since the chemical property prediction is typically evaluated via AUC~\cite{wu2018moleculenet}, we define a robust version of AUC that takes account of the radius of the adversary as well as the ratio of testing data that can be manipulated.  Note that to maximally decrease the score of AUC via a positive (negative) example, the adversary only has to maximally decrease (increase) its prediction probability, regardless of the scores of the other examples. Hence, given an $\ell_0$ radius $r$ and a ratio of testing data, we first compute the adversary for each testing data, and then find the combination of adversaries and the clean data under the ratio constraint that leads to the worst AUC score. \textcyan{See details in Appendix~\ref{sec:qp}.}

The results are in Figure~\ref{fig:bace}. Empirically, the adversary of the decision tree at $r = 1$ always changes the prediction probability of a positive (negative) example to $0$ ($1$). Hence, the plots of the decision tree model are constant across different $\ell_0$ radii. The randomly smoothed decision tree is consistently more robust than the vanilla decision tree model. We also compare the exact certificate of the prediction probability with the one derived from Lemma~\ref{thm:main}; the average difference across the training data is $0.358$ and $0.402$ when $r$ equals to $1$ and $2$, respectively.  The phenomenon encourages the development of a classifier-aware guarantee that is tighter than the classifier-agnostic guarantee.

\section{Conclusion}
We present a stratified approach to certifying the robustness of randomly smoothed classifiers, where the robustness guarantees can be obtained in various resolutions and perspectives, ranging from a point-wise certificate to a regional certificate and from general results to specific examples. The hierarchical investigation opens up many avenues for future extensions at different levels. 
\subsubsection*{Acknowledgments}
GH and TJ were in part supported by a grant from Siemens Corporation.

\bibliographystyle{abbrv}
\bibliography{reference}

\begin{thebibliography}{10}

\bibitem{bemis1996properties}
G.~W. Bemis and M.~A. Murcko.
\newblock The properties of known drugs. 1. molecular frameworks.
\newblock {\em Journal of medicinal chemistry}, 39(15):2887--2893, 1996.

\bibitem{cao2017mitigating}
X.~Cao and N.~Z. Gong.
\newblock Mitigating evasion attacks to deep neural networks via region-based
  classification.
\newblock In {\em Proceedings of the 33rd Annual Computer Security Applications
  Conference}, pages 278--287. ACM, 2017.

\bibitem{carlini2017provably}
N.~Carlini, G.~Katz, C.~Barrett, and D.~L. Dill.
\newblock Provably minimally-distorted adversarial examples.
\newblock {\em arXiv preprint arXiv:1709.10207}, 2017.

\bibitem{cheng2017maximum}
C.-H. Cheng, G.~N{\"u}hrenberg, and H.~Ruess.
\newblock Maximum resilience of artificial neural networks.
\newblock In {\em International Symposium on Automated Technology for
  Verification and Analysis}, pages 251--268. Springer, 2017.

\bibitem{clopper1934use}
C.~J. Clopper and E.~S. Pearson.
\newblock The use of confidence or fiducial limits illustrated in the case of
  the binomial.
\newblock {\em Biometrika}, 26(4):404--413, 1934.

\bibitem{cohen2019certified}
J.~M. Cohen, E.~Rosenfeld, and J.~Z. Kolter.
\newblock Certified adversarial robustness via randomized smoothing.
\newblock In {\em the 36th International Conference on Machine Learning}, 2019.

\bibitem{croce2018provable}
F.~Croce, M.~Andriushchenko, and M.~Hein.
\newblock Provable robustness of relu networks via maximization of linear
  regions.
\newblock In {\em the 22nd International Conference on Artificial Intelligence
  and Statistics}, 2018.

\bibitem{deng2009imagenet}
J.~Deng, W.~Dong, R.~Socher, L.-J. Li, K.~Li, and L.~Fei-Fei.
\newblock Imagenet: A large-scale hierarchical image database.
\newblock In {\em Proceedings of the IEEE international conference on computer
  vision}, pages 248--255. Ieee, 2009.

\bibitem{dutta2018output}
S.~Dutta, S.~Jha, S.~Sankaranarayanan, and A.~Tiwari.
\newblock Output range analysis for deep feedforward neural networks.
\newblock In {\em NASA Formal Methods Symposium}, pages 121--138. Springer,
  2018.

\bibitem{dvijotham2018training}
K.~Dvijotham, S.~Gowal, R.~Stanforth, R.~Arandjelovic, B.~O'Donoghue,
  J.~Uesato, and P.~Kohli.
\newblock Training verified learners with learned verifiers.
\newblock {\em arXiv preprint arXiv:1805.10265}, 2018.

\bibitem{dvijotham2018dual}
K.~Dvijotham, R.~Stanforth, S.~Gowal, T.~Mann, and P.~Kohli.
\newblock A dual approach to scalable verification of deep networks.
\newblock In {\em the 34th Annual Conference on Uncertainty in Artificial
  Intelligence}, 2018.

\bibitem{ehlers2017formal}
R.~Ehlers.
\newblock Formal verification of piece-wise linear feed-forward neural
  networks.
\newblock In {\em International Symposium on Automated Technology for
  Verification and Analysis}, pages 269--286. Springer, 2017.

\bibitem{finlay2019logbarrier}
C.~Finlay, A.-A. Pooladian, and A.~M. Oberman.
\newblock The logbarrier adversarial attack: making effective use of decision
  boundary information.
\newblock {\em arXiv preprint arXiv:1903.10396}, 2019.

\bibitem{fischetti2018deep}
M.~Fischetti and J.~Jo.
\newblock Deep neural networks and mixed integer linear optimization.
\newblock {\em Constraints}, 23:296--309, 2018.

\bibitem{goodfellow14}
I.~Goodfellow, J.~Shlens, and C.~Szegedy.
\newblock Explaining and harnessing adversarial examples.
\newblock In {\em International Conference on Learning Representations}, 2015.

\bibitem{gowal2018effectiveness}
S.~Gowal, K.~Dvijotham, R.~Stanforth, R.~Bunel, C.~Qin, J.~Uesato, T.~Mann, and
  P.~Kohli.
\newblock On the effectiveness of interval bound propagation for training
  verifiably robust models.
\newblock {\em arXiv preprint arXiv:1810.12715}, 2018.

\bibitem{he2016deep}
K.~He, X.~Zhang, S.~Ren, and J.~Sun.
\newblock Deep residual learning for image recognition.
\newblock In {\em Proceedings of the IEEE conference on computer vision and
  pattern recognition}, pages 770--778, 2016.

\bibitem{huang2019achieving}
P.-S. Huang, R.~Stanforth, J.~Welbl, C.~Dyer, D.~Yogatama, S.~Gowal,
  K.~Dvijotham, and P.~Kohli.
\newblock Achieving verified robustness to symbol substitutions via interval
  bound propagation.
\newblock {\em arXiv preprint arXiv:1909.01492}, 2019.

\bibitem{jia2019certified}
R.~Jia, A.~Raghunathan, K.~G{\"o}ksel, and P.~Liang.
\newblock Certified robustness to adversarial word substitutions.
\newblock {\em arXiv preprint arXiv:1909.00986}, 2019.

\bibitem{katz2017reluplex}
G.~Katz, C.~Barrett, D.~L. Dill, K.~Julian, and M.~J. Kochenderfer.
\newblock Reluplex: An efficient smt solver for verifying deep neural networks.
\newblock In {\em International Conference on Computer Aided Verification},
  pages 97--117. Springer, 2017.

\bibitem{lecuyer2018certified}
M.~Lecuyer, V.~Atlidakis, R.~Geambasu, D.~Hsu, and S.~Jana.
\newblock Certified robustness to adversarial examples with differential
  privacy.
\newblock {\em IEEE Symposium on Security and Privacy (SP)}, 2019.

\bibitem{lee2018towards}
G.-H. Lee, D.~Alvarez-Melis, and T.~S. Jaakkola.
\newblock Towards robust, locally linear deep networks.
\newblock In {\em International Conference on Learning Representations}, 2019.

\bibitem{li2018second}
B.~Li, C.~Chen, W.~Wang, and L.~Carin.
\newblock Second-order adversarial attack and certifiable robustness.
\newblock {\em arXiv preprint arXiv:1809.03113}, 2018.

\bibitem{liu2018towards}
X.~Liu, M.~Cheng, H.~Zhang, and C.-J. Hsieh.
\newblock Towards robust neural networks via random self-ensemble.
\newblock In {\em Proceedings of the European Conference on Computer Vision
  (ECCV)}, pages 369--385, 2018.

\bibitem{lomuscio2017approach}
A.~Lomuscio and L.~Maganti.
\newblock An approach to reachability analysis for feed-forward relu neural
  networks.
\newblock {\em arXiv preprint arXiv:1706.07351}, 2017.

\bibitem{madry2017towards}
A.~Madry, A.~Makelov, L.~Schmidt, D.~Tsipras, and A.~Vladu.
\newblock Towards deep learning models resistant to adversarial attacks.
\newblock In {\em International Conference on Learning Representations}, 2018.

\bibitem{mirman2018differentiable}
M.~Mirman, T.~Gehr, and M.~Vechev.
\newblock Differentiable abstract interpretation for provably robust neural
  networks.
\newblock In {\em the 35th International Conference on Machine Learning}, 2018.

\bibitem{neyman1933ix}
J.~Neyman and E.~S. Pearson.
\newblock Ix. on the problem of the most efficient tests of statistical
  hypotheses.
\newblock {\em Philosophical Transactions of the Royal Society of London.
  Series A, Containing Papers of a Mathematical or Physical Character},
  231(694-706):289--337, 1933.

\bibitem{paszke2017automatic}
A.~Paszke, S.~Gross, S.~Chintala, G.~Chanan, E.~Yang, Z.~DeVito, Z.~Lin,
  A.~Desmaison, L.~Antiga, and A.~Lerer.
\newblock Automatic differentiation in pytorch.
\newblock 2017.

\bibitem{raghunathan2018certified}
A.~Raghunathan, J.~Steinhardt, and P.~Liang.
\newblock Certified defenses against adversarial examples.
\newblock In {\em International Conference on Learning Representations}, 2018.

\bibitem{raghunathan2018semidefinite}
A.~Raghunathan, J.~Steinhardt, and P.~S. Liang.
\newblock Semidefinite relaxations for certifying robustness to adversarial
  examples.
\newblock In {\em Advances in Neural Information Processing Systems}, pages
  10877--10887, 2018.

\bibitem{rogers2010extended}
D.~Rogers and M.~Hahn.
\newblock Extended-connectivity fingerprints.
\newblock {\em Journal of chemical information and modeling}, 50(5):742--754,
  2010.

\bibitem{scheibler2015towards}
K.~Scheibler, L.~Winterer, R.~Wimmer, and B.~Becker.
\newblock Towards verification of artificial neural networks.
\newblock In {\em MBMV}, pages 30--40, 2015.

\bibitem{singh2018fast}
G.~Singh, T.~Gehr, M.~Mirman, M.~P{\"u}schel, and M.~Vechev.
\newblock Fast and effective robustness certification.
\newblock In {\em Advances in Neural Information Processing Systems}, pages
  10802--10813, 2018.

\bibitem{subramanian2016computational}
G.~Subramanian, B.~Ramsundar, V.~Pande, and R.~A. Denny.
\newblock Computational modeling of $\beta$-secretase 1 (bace-1) inhibitors
  using ligand based approaches.
\newblock {\em Journal of chemical information and modeling},
  56(10):1936--1949, 2016.

\bibitem{tjeng2017evaluating}
V.~Tjeng, K.~Xiao, and R.~Tedrake.
\newblock Evaluating robustness of neural networks with mixed integer
  programming.
\newblock In {\em International Conference on Learning Representations}, 2017.

\bibitem{tocher1950extension}
K.~Tocher.
\newblock Extension of the neyman-pearson theory of tests to discontinuous
  variates.
\newblock {\em Biometrika}, 37(1/2):130--144, 1950.

\bibitem{weng2018towards}
T.-W. Weng, H.~Zhang, H.~Chen, Z.~Song, C.-J. Hsieh, D.~Boning, I.~S. Dhillon,
  and L.~Daniel.
\newblock Towards fast computation of certified robustness for relu networks.
\newblock In {\em the 35th International Conference on Machine Learning}, 2018.

\bibitem{wong2017provable}
E.~Wong and J.~Z. Kolter.
\newblock Provable defenses against adversarial examples via the convex outer
  adversarial polytope.
\newblock In {\em the 35th International Conference on Machine Learning}, 2018.

\bibitem{wong2018scaling}
E.~Wong, F.~Schmidt, J.~H. Metzen, and J.~Z. Kolter.
\newblock Scaling provable adversarial defenses.
\newblock In {\em Advances in Neural Information Processing Systems}, pages
  8400--8409, 2018.

\bibitem{wu2018moleculenet}
Z.~Wu, B.~Ramsundar, E.~N. Feinberg, J.~Gomes, C.~Geniesse, A.~S. Pappu,
  K.~Leswing, and V.~Pande.
\newblock Moleculenet: a benchmark for molecular machine learning.
\newblock {\em Chemical science}, 9(2):513--530, 2018.

\bibitem{zhang2018efficient}
H.~Zhang, T.-W. Weng, P.-Y. Chen, C.-J. Hsieh, and L.~Daniel.
\newblock Efficient neural network robustness certification with general
  activation functions.
\newblock In {\em Advances in Neural Information Processing Systems}, pages
  4939--4948, 2018.

\end{thebibliography}

\clearpage
\newpage
\appendix
\section{Proofs}
\label{sec:proofs}

To simplify exposition, we use $[n]$ to denote the set $\{1, 2, \dots, n\}$.

\subsection{The proof of Proposition~\ref{prop:uniform}}


\begin{proof}
We have
\begin{align*}
\vspace{-1.5mm}
    \begin{cases}
    \rho_{\vx, \bar \vx}(p)  = 0, &\text{ if } 0 \leq p \leq \Pr(\phi(\vx) \in \gL_1),\\
    \rho_{\vx, \bar \vx}(p)  = p - \Pr(\phi(\vx) \in \gL_1), &\text{ if } 1 \geq p > \Pr(\phi(\vx) \in \gL_1),
    \end{cases}
\vspace{-0.5mm}
\end{align*}
where $\Pr(\phi(\vx) \in \gL_1) = \text{Vol}(\gB_1 \backslash \gB_2) / \text{Vol}(\gB_1)$ and $\text{Vol}(\gB_1)$ is a constant given $\gamma$.
Hence, the minimizers of $\min_{\bar \vx \in \gB_{r, q}(\vx)} \rho_{\vx, \bar \vx}(p)$ are simply the points that maximize the volume of $\gB_1 \backslash \gB_2$, or, equivalently, minimize the volume of $\gB_1 \cap \gB_2$.
Below we re-write $\bar \vx$ as $\vx + \vdel$.

\emph{Case $q=1$:}
 $\forall r > 0$, we want to find a $\vdel$ s.t., $\|\vdel\|_1=r$ and the overlapping region is minimized: (By symmetry, we  assume that $\vdel_i\geq 0$ for all $i$)
 \begin{align}
     \argmin_{\vdel\geq \mathrm{0}: \|\vdel \|_1 = r} \prod_{i=1}^d (2\gamma - \vdel_i).
     \label{eq:l1}
 \end{align}
 
 Since $\forall i, j \in [d], i\neq j$, we know
 \begin{align}
     (2\gamma - \vdel_i)(2\gamma - \vdel_j) = 4\gamma^2 - (\vdel_i + \vdel_j) + \vdel_i \vdel_j \geq 4\gamma^2 - (\vdel_i + \vdel_j).
 \end{align}

So we can always move the mass of $\vdel_j$ to $\vdel_i$ to further decrease the product value. That means, $\vdel_1 = r, \vdel_i = 0, \forall i \neq 1$ minimizes Eq. (\ref{eq:l1}) for a given $r$. As a result, we know 
 \begin{align}
     & \sup r, s.t.\text{   } \min_{\vdel: \|\vdel\|_1 \leq r} \rho_{\vx,  \vx+\vdel}(p) > 0.5 \\
     = & \sup r, s.t.\text{   } p - \left(1 - \frac{(2\gamma)^{d-1}(2\gamma - r)}{(2\gamma)^{d}}\right) > 0.5 \\
     = & 2p \gamma - \gamma
 \end{align}

\emph{Case $q=\infty$:}
Similarly, for $q=\infty$ case, we want to find a $\vdel$ with $\|\vdel\|_\infty=r$, and the following is minimized: (by symmetry we assume $\vdel_i\geq 0$ for all $i$)
 \begin{align*}
     \argmin_{\vdel\geq \mathrm{0}: \|\vdel \|_\infty  = r} \prod_{i=1}^d (2\gamma - \vdel_i).
 \end{align*}
 
 In this case, we should set $\vdel_i=r$ for all $i$, which means  

 \begin{align}
     & \sup r, s.t.\text{   } \min_{\vdel: \|\vdel\|_\infty \leq r} \rho_{\vx,  \vx+\vdel}(p) > 0.5 \\
     = & \sup r, s.t.\text{   }  p - \left(1 - \frac{(2\gamma - r)^d}{(2\gamma)^d}\right) > 0.5
     \\
     = & \sup r, s.t.\text{   }    \frac{(2\gamma - r)^d}{(2\gamma)^d}> 1.5-p
 \end{align}
It remains to see that
 \begin{align*}
     & \frac{(2\gamma - r)^d}{(2\gamma)^d} > 1.5 - p \\
     \iff & {2\gamma - r} > 2\gamma(1.5 - p)^{1/d} \\
     \iff & {2\gamma - } 2\gamma(1.5 - p)^{1/d} > r. \qedhere
 \end{align*}
\end{proof}

\subsection{The proof of Lemma~\ref{thm:main}}


\begin{proof}
$\forall \bar f \in \gF$, We may rewrite the probabilities in an integral form:
\begin{align*}
    \Pr(\bar f(\rand(\vx)) = y)      & = \sum_{i=1}^n \int_{\gL_i} \Pr(\rand(\vx) = \vz) \Pr(\bar f(\vz) = y)  \mathrm{d}\vz, \\
    \Pr(\bar f(\rand(\bar \vx)) = y) & = \sum_{i=1}^n \int_{\gL_i} \Pr(\rand(\bar \vx) = \vz) \Pr(\bar f(\vz) = y)  \mathrm{d}\vz 
\end{align*}
Note that for all possible $\bar f \in \gF$, we can re-assign all the function output within a likelihood region to be \emph{constant} without affecting $\Pr(\bar f(\rand(\vx)) = y)$ and $\Pr(\bar f(\rand(\bar \vx)) = y)$. Concretely, we define $\bar f'$ as
\begin{align*}
    \Pr(\bar f'(\vz') = y) = \frac{\int_{\gL_i} \Pr(\rand(\vx) = \vz) \Pr(\bar f(\vz) = y) \mathrm{d}\vz}{\int_{\gL_i} \Pr(\rand(\vx) = \vz) \mathrm{d}\vz}, \forall \vz' \in \gL_i, \forall i \in [n],
\end{align*}
then we have
\[
\int_{\gL_i} \Pr(\rand(\vx) = \vz) \Pr(\bar f(\vz) = y) \mathrm{d}\vz
=
\int_{\gL_i} \Pr(\rand(\vx) = \vz) \Pr(\bar f'(\vz) = y) \mathrm{d}\vz
\]
Since in $\gL_i$,  
$\Pr(\rand(\vx) = \vz)/\Pr(\rand(\bar \vx) = \vz)$ is constant, we also have
\[
\int_{\gL_i} \Pr(\rand(\bar \vx) = \vz) \Pr(\bar f(\vz) = y) \mathrm{d}\vz
=
\int_{\gL_i} \Pr(\rand(\bar \vx) = \vz) \Pr(\bar f'(\vz) = y) \mathrm{d}\vz
\]
Therefore, 
\begin{align*}
&    \Pr(\bar f(\rand(\vx)) = y) = \Pr(\bar f'(\rand(\vx)) = y), \text{ and}\\
&        \Pr(\bar f(\rand(\bar \vx)) = y) = \Pr(\bar f'(\rand(\bar \vx)) = y) .
\end{align*}

Hence, it suffices to consider the following program
\begin{align*}
    \mathrm{(I)} \triangleq \min_{g: [n] \to [0, 1]} & \sum_{i=1}^n \int_{\gL_i} \Pr(\rand(\bar \vx) = \vz) g(i) \mathrm{d}\vz,\\
    s.t.\;\; & \sum_{i=1}^n \int_{\gL_i} \Pr(\rand(\vx) = \vz) g(i) \mathrm{d}\vz = p,
\end{align*}
where the optimum is equivalent to the program
\begin{align*}
    \min_{\bar f \in \gF: \Pr(\bar f(\rand(\vx)) = y) = p} & \Pr(\bar f(\rand(\bar \vx)) =  y),
\end{align*}
and the each $g$ corresponds to a solution $\bar f$. 
For example, the $f^*$ in the statement corresponds to the $g^*$ defined as:
\begin{align}
    g^*(i) =\begin{cases}
         1,  &\text{if } i < H^*,\\
        \frac{p - \sum_{i=1}^{H^*-1} \Pr(\rand(\vx) \in \gL_i)}{\Pr(\rand(\vx) \in \gL_{H^*})},  &\text{if } i = H^*,\\
         0, &\text{if } i > H^*.
        \end{cases}
\end{align}

We may simplify the program as
\begin{align*}
    \mathrm{(I)} = \min_{g: [n] \to [0, 1]} & \sum_{i=1}^n \Pr(\rand(\bar \vx) \in \gL_i) g(i),\\
    s.t.\;\; & \sum_{i=1}^n \Pr(\rand(\vx)\in \gL_i) g(i) = p.
\end{align*}
Clearly, if $\eta_i = 0$, all the optimal $g$ will assign $g(i) = 0$; our solution $g^*$ satisfies this property since
\begin{align}
H^* \triangleq \min_{H \in \{1,\dots,n\}: \sum_{i=1}^H \Pr(\rand(\vx) \in \gL_i) \geq p} H
\end{align}
implies $\eta_{H^*} > 0$ (otherwise, it implies that $\Pr(\rand(\vx) \in \gL_{H^*}) = 0$ and leads to a contradiction).
Hence, we can ignore the regions with $\eta_i = 0$, assume $\eta_n > 0$, and simplify program $\mathrm{(I)}$ again as
\begin{align}
    \mathrm{(I)} = \min_{g: [n] \to [0, 1]} & \sum_{i=1}^n \frac{1}{\eta_i} \Pr(\rand(\vx) \in \gL_i) g(\eta_i), \label{eq:p1:obj}\\
    s.t.\;\; & \sum_{i=1}^n \Pr(\rand(\vx) \in \gL_i) g(\eta_i) = p. \label{eq:p1:constraint}
\end{align}

It is evident that $g^*$ satisfies the constraint (\ref{eq:p1:constraint}), and we will prove that any $g \neq g^*$ that satisfies constraint (\ref{eq:p1:constraint}) cannot be better.

$\forall g: [n] \to [0, 1]$, we define $\Delta(i) \triangleq (g^*(i) - g(i)) P(\rand(\vx) \in \gL_i)$. Then we have
\begin{align}
      & \sum_{i=1}^n \frac{1}{\eta_i} \Pr(\rand(\vx) \in \gL_i) g(i) = \sum_{i=1}^n \frac{1}{\eta_i} \bigg[ \Pr(\rand(\vx) \in \gL_i) g^*(i) - \Delta(i) \bigg] \nonumber\\
    = & \sum_{i=1}^{H^*} \frac{1}{\eta_i} \Pr(\rand(\vx) \in \gL_i) g^*(i) - \sum_{i=1}^n \frac{1}{\eta_i} \Delta(i). \label{eq:discrete-deriv}
\end{align}
Note that $\Delta(i) \geq 0$ for $i < H^*$, $\Delta(i) \leq 0$ for $i > H^*$, and $\sum_{i=1}^n \Delta(\eta_i) = 0$ due to the constraint (\ref{eq:p1:constraint}). Therefore, we have
\begin{align}
    \sum_{i=1}^n \frac{1}{\eta_i} \Delta(\eta_i) \leq \sum_{i=1}^{n} \frac{1}{\eta_{H^*}} \Delta(\eta_i) = 0. \label{eq:discrete-ineq}
\end{align}
Finally, combining (\ref{eq:discrete-deriv}) and (\ref{eq:discrete-ineq}), 
\begin{align}
      & \sum_{i=1}^n \frac{1}{\eta_i} \Pr(\rand(\vx) \in \gL_i) g(i) \geq \sum_{i=1}^{H^*} \frac{1}{\eta_i} \Pr(\rand(\vx) \in \gL_i) g^*(i). 
\end{align}
\end{proof}

\subsection{The proof of Lemma~\ref{lem:canonical}}
\label{appendix:proof:canonical}

\begin{proof}
If $\baserand(\cdot)$ is defined as Eq.~(\ref{eqn:perturb}), $\forall \vx, \bar \vx \in \gX$ such that $ \|\vx- \bar \vx\|_0=r$, below we show that $\rho_{\vx, \bar \vx}$ is independent of $\vx$ and $\bar \vx$. 
Indeed, since $\|\vx- \bar \vx\|_0$ is the number of non-zero elements of $\vx- \bar \vx$, we know there are exactly $r$ dimensions such that $\vx$ and $\bar \vx$ do not match. 
Notice that $\baserand(\cdot)$ applies to each dimension of $\vx$ independently, so we can safely ignore any correlations between two dimensions. 
Therefore, by the symmetry of the distribution, we can rearrange the \emph{order} of coordinates, and assume $\vx$ and $\bar \vx$ differ for the first $r$ dimensions, and match for the rest $d-r$ dimensions. 

Notice that the randomization $\baserand(\cdot)$ has the nice property that the perturbing probabilities are oblivious to the \emph{actual values} of the input. 
Therefore, by the definition of $\vx_C, \bar \vx_C$,
we know that they are the canonical form of all pairs of $\vx$ and $\bar \vx$ such that $\|\vx-\bar \vx\|_0=r$; hence, $\rho_{\vx, \bar \vx}(p)$ is constant and equals $\rho_r(p)$ for every $p \in [0, 1]$.
\end{proof}

\subsection{The proof of Lemma~\ref{lem:computation}}


\begin{proof}
%
In this proof, we adopt the notation of the canonical form $\vx_C$ and $\bar \vx_C$ from Appendix~\ref{appendix:proof:canonical}.

Recall that $\vx_C$ is a zero vector, and $\bar \vx_C$ has the first $r$ entries equal to $1$ and the last $d-r$ entries equal to $0$. We use the likelihood tuple $(u, v)$ to refer the scenario when $\vx_C$ ``flips'' $u$ coordinates (the likelihood is $\alpha^{d-u} \beta^{u}$), and $\bar \vx_C$ ``flips'' $v$ coordinates (the likelihood is $\alpha^{d-v} \beta^{v}$). 
Note that $u \leq v$ by assumption.
For $r \in [d]$ and $(u, v) \in \{0, 1, \dots, d\}^2$, the number of possible outcome $\vz \in \gX$ with the likelihood tuple $(u, v)$ can be computed in the following way:
\begin{align*}
    |\gL(u, v; r)| = \sum_{i=0}^{\min(u, d-r)} \sum_{j=0}^{u-i}  \frac{(K - 1)^j \one((u-i-j) + (v-i-j) + j = r) r !}{(u-i-j)!(v-i-j)!j!} \frac{K^i(d - r)!}{(d-r-i)! i!},
\end{align*}
where the first summation and the term 
\begin{equation*}
\frac{K^i(d - r)!}{(d-r-i)! i!} 
\end{equation*}
correspond to the case where $i$ entries out of the last $(d-r)$ coordinates in $\vx_C$ and $\bar \vx_C$ are both modified. Notice that $\vx_C$ and $\bar \vx_C$ are equal in the last $d-r$ dimensions, so if $\vx_C$ has $i$ entries modified among them, in order to ensure that $\bar \vx_C$ equals $\vx_C$ after modification, $\bar \vx_C$ should have exactly the same $i$ entries modified as well (in order to become the same $\vz$ in Eq.~(\ref{eqn:LMN})).

The second summation and the term 
\begin{equation*}
    \frac{(K - 1)^j \one((u-i-j) + (v-i-j) + j = r) r !}{(u-i-j)!(v-i-j)!j!}
\end{equation*}
corresponds to the case where $j$ entries out of the first $r$ coordinates of $\vx_C$ and $\bar \vx_C$ are modified to any values other than $\{0, 1\}$, $u - i - j$ entries in $\vx_C$ are modified to $1$, and $v - i - j$ entries in $\bar \vx_C$ are modified to $0$. By the same analysis, we know that both $\vx_C$ and $\bar \vx_C$ should have exactly the same $j$ entries modified to any value other than $\{0, 1\}$. 
The indicator function $\one((u-i-j) + (v-i-j) + j = r) $ simply verifies that whether the value of $j$ is valid. 
Note that these two summations have covered all possible cases of modifications on $\vx_C$ and $\bar \vx_C$ in $\gL(u, v; r)$.

After fixing the value of $i$ and $j$, each summand is simply calculating the number of symmetric cases. $(K-1)^j$ means there are $j$ entries modified to $(K-1)$ possible values. $\frac{r!}{(u-i-j)!(v-i-j)!j!}$ is the number of possible configurations for the first $r$ coordinates. $K^i$ means there are $i$ entries momdified to $K$ possible values. $\frac{(d-r)!}{(d-r-1)!i!}$ is the number of possible configurations for the last $d-r$ coordinates. 

Now it remains to simplify the expression. Let $j^* \triangleq u + v - 2i - r$, we have
\begin{align}
    |\gL(u, v; r)|& =  \sum_{i=0}^{\min(u, d-r)} \frac{\one(j^* \geq 0) \one(j^* \leq u-i) (K - 1)^{j^*} r !}{(u-i-j^*)!(v-i-j^*)!j^*!} \frac{K^i(d - r)!}{(d-r-i)! i!},  \nonumber\\
    & = \sum_{i=\max\{0, v - r\}}^{\min(u, d-r)} \frac{\one(j^* \geq 0) (K - 1)^{j^*} r !}{(u-i-j^*)!(v-i-j^*)!j^*!} \frac{K^i(d - r)!}{(d-r-i)! i!},  \nonumber\\
    & = \sum_{i=\max\{0, v - r\}}^{\min(u, d-r, \lfloor \frac{u+v-r}{2} \rfloor )} \frac{(K - 1)^{j^*} r !}{(u-i-j^*)!(v-i-j^*)!j^*!} \frac{K^i(d - r)!}{(d-r-i)! i!}. \label{eq:compute:2nd}
\end{align}
Moreover, we know that $|\gL(u, v; r)| = |\gL(v, u; r)|$ holds by the symmetry between $\vx_C$ and $\bar \vx_C$.
\end{proof}

\section{Algorithms For Decision Tree}

\subsection{Dynamic Programming For Restricted Decision Tree}
\label{sec:dp_tree}

Given an input $\vx\in \gX$, 
we run dynamic programming (Algorithm \ref{alg:dp}) for computing the certificate, based on the same idea mentioned in Section~\ref{sec:refinement}.

\begin{algorithm}[H]
\begin{algorithmic}[1]
\IF {$i$ is leaf}
\FOR {$r=1, \cdots, R$}
   \STATE $\texttt{adv}[i,r]=$Leaf-Output$(i)$
 \ENDFOR
 \STATE Return
\ENDIF
\STATE $rw=\alpha^{\one\{\vx_{\texttt{idx}[i]} = 1\}}\beta^{\one\{\vx_{\texttt{idx}[i]} = 0\}}$
\STATE $lw=\alpha^{\one\{\vx_{\texttt{idx}[i]} = 0\}}\beta^{\one\{\vx_{\texttt{idx}[i]} = 1\}}$
\STATE DP($\vx$, \texttt{right}[i], $R$)
\STATE DP($\vx$, \texttt{left}[i], $R$)
\FOR {$r=1, \cdots, R$}
\STATE $\texttt{adv}[i,r] =1$
\FOR {$\bar r=0, \cdots r$}
\STATE
$\texttt{adv}[i,r]=\min\{
\texttt{adv}[i,r],
rw*\texttt{adv}[\texttt{right}[i],\bar r]
+
lw*\texttt{adv}[\texttt{left}[i],r-\bar r]
\}$
\ENDFOR 
\FOR {$\bar r=0, \cdots r-1$}
\STATE
$\texttt{adv}[i,r]=\min\{
\texttt{adv}[i,r],
lw*\texttt{adv}[\texttt{right}[i],\bar r]
+
rw*\texttt{adv}[\texttt{left}[i],r-1-\bar r]
\}$
\ENDFOR 
\ENDFOR

 \end{algorithmic}
\caption{DP($\vx$, $i$, $R$)}
\label{alg:dp}
\end{algorithm}

We use $\texttt{adv}[i,r]$ to denote the worst prediction at node $i$ if at most $r$ features can be perturbed. 
Algorithm \ref{alg:dp} uses the following updating rule
for $\texttt{adv}[i,r]$.
\begin{align}
    &\texttt{adv}[i, r] = \min\{  \nonumber\\
    &\min_{\bar r \in \{0, 1, \dots, r\} }\{ \alpha^{\one\{\vx_{\texttt{idx}[i]} = 1\}}\beta^{\one\{\vx_{\texttt{idx}[i]} = 0\}} \texttt{adv}[\texttt{right}[i], \bar r] + \alpha^{\one\{\vx_{\texttt{idx}[i]} = 0\}}\beta^{\one\{\vx_{\texttt{idx}[i]} = 1\}} \texttt{adv}[\texttt{left}[i], r - \bar r] \}, \nonumber\\
    &\min_{\bar r \in \{0, 1, \dots, r-1\} }\{ \alpha^{\one\{\vx_{\texttt{idx}[i]} = 0\}}\beta^{\one\{\vx_{\texttt{idx}[i]} = 1\}} \texttt{adv}[\texttt{right}[i], \bar r] + \alpha^{\one\{\vx_{\texttt{idx}[i]} = 1\}}\beta^{\one\{\vx_{\texttt{idx}[i]} = 0\}} \texttt{adv}[\texttt{left}[i], r - 1 - \bar r] \} \}\nonumber
\end{align}

There are two cases in this updating rule. In the first case, the feature used at node $i$ is not perturbed, so it remains to see
if we perturb $\bar r$ features in the right subtree and $r-\bar r$ features in the left subtree, what is the minimum adversarial prediction if $\bar r \in \{0, \cdots, r\}$. In the second case, the feature used at node $i$ is perturbed, and we check if we perturb $r-1$ features in the two subtrees, what is the minimum adversarial prediction. Combining the two cases together, we get the solution for $\texttt{adv}[i, r]$.





\subsection{Training Algorithm For Decision Tree}
\label{sec:tree_training}

We consider the randomization scheme introduced in Eq.~(\ref{eqn:perturb}): for every coordinate in a given input $\vx$, we may perturb its value with probability $\beta$. After perturbation, $\vx$ may arrive at any leaf node, rather than following one specific path as in the standard decision tree. 
Therefore, when training, we 
maintain the probability of arriving at the current tree node for every input $\vx$, denoted as $\texttt{probs}$. The probability is multiplied by $\alpha$ or $\beta$ after each layer, depending on the input and the feature used for the current tree node. See Algorithm \ref{alg:soft_tree} for details.


\begin{figure}
    \centering
\begin{algorithm}[H]
\begin{algorithmic}[1]
\STATE $Q=[(\texttt{root},0, [1,1,\cdots, 1])]$
\WHILE{Q not Empty}
\STATE $i, \texttt{dep}, \texttt{probs}=Q.\texttt{pop}()$
\IF {\texttt{dep}=\texttt{maxdep}}
\STATE Assign-Leaf-Node$(i, \texttt{probs})$
\STATE Continue
\ENDIF
\STATE f-list = Get-available-features$()$
\FOR {\texttt{idx} in f-list}
\STATE $\texttt{list}=[(y, \alpha^{\one\{\vx_{\texttt{idx}} = 1\}}\beta^{\one\{\vx_{\texttt{idx}} = 0\}}\texttt{probs}[\vx])$ for $(\vx,y)$ in $X$]
\STATE $\texttt{idx}^*=$ Update-best-feature-score
$(\texttt{idx},\texttt{list},\texttt{idx}^*)$
\ENDFOR
\STATE \texttt{idx}[$i$]$=\texttt{idx}^*$
\STATE \texttt{left-probs}=\texttt{probs}
\STATE \texttt{right-probs}=\texttt{probs}
\FOR {$\vx$ in $X$}
\IF {$\vx_{\texttt{idx}}=1$}
\STATE \texttt{left-probs}$[\vx]=$\texttt{left-probs}$[\vx]*\alpha $
\STATE \texttt{right-probs}$[\vx]=$\texttt{right-probs}$[\vx]*\beta$
\ELSE
\STATE \texttt{right-probs}$[\vx]=$\texttt{right-probs}$[\vx]*\alpha $
\STATE \texttt{left-probs}$[\vx]=$\texttt{left-probs}$[\vx]*\beta$
\ENDIF
\ENDFOR
\STATE $Q.\texttt{push}(\texttt{left}[i], \texttt{dep}+1, \texttt{left-probs})$
\STATE $Q.\texttt{push}(\texttt{right}[i], \texttt{dep}+1, \texttt{right-probs})$
\ENDWHILE
 \end{algorithmic}
\caption{Train($X$,$Y$, \texttt{maxdep})}
\label{alg:soft_tree}
\end{algorithm}
\end{figure}

The overall framework of Algorithm \ref{alg:soft_tree} is standard: we train the tree nodes greedily in breadth-first ordering, and pick the best splitting feature every time. 
However, when picking the best splitting feature, 
the standard decision tree uses Gini impurity based on  all the remaining training data that will follow the path from the root to the current tree node. 
In our algorithm, this will  include all the training data, but with different arriving probabilities.   Therefore, we apply the weighted Gini impurity metric instead. Specifically, for a split, its weighted Gini impurity is (after \texttt{probs} is updated with \texttt{idx}):
\[
1- \left(\frac{\sum_{\vx\in \gX, y = 1}\texttt{probs}[\vx]}{\sum_{\vx\in \gX}\texttt{probs}[\vx]}\right)^2-
\left(\frac{\sum_{\vx\in \gX, y = 0}\texttt{probs}[\vx]}{\sum_{\vx\in \gX}\texttt{probs}[\vx]}\right)^2
\]

When the arriving probability for each $\vx$ is restricted to be either $0$ or $1$, this definition becomes the standard  Gini impurity.

\section{Experimental Details}
\subsection{Supplementary Materials for the MNIST Experiment}
\label{sec:CNN_model}

For the MNIST experiment, we use a simple 4-layer convolutional network, where the first two layers are convolutional layers, and the last two layers are feedforward layers. For the two convolutional layers, we use kernel size $5$, and output channels $20$ and $50$, respectively. For the two feedforward layers, we use $500$ hidden nodes.
We tune the hyperparameter  $\alpha \in \{0.72,  0.76, 0.80, 0.84, 0.88, 0.92, 0.96\}$ for $\mu(R)$ in the validation set using the certificates from the Gaussian distribution~\cite{cohen2019certified}. The resulting $\alpha$ is $0.8$. 
\textred{
We also train a CNN with an isotropic Gaussian with the $\sigma$ that corresponds to $\alpha=0.8$ (see \xref{sec:comparison}).}

The learning procedure for all the models are the same (except the distribution). The batch size is $400$. We train each model for $30$ epochs with the SGD optimizer with Nesterov momentum (momentum = $0.9$). The learning rate is initially set to be $0.05$ and annealed by a factor of 10 for every $10$ epochs of training. The models are implemented in \texttt{PyTorch}~\cite{paszke2017automatic}, and run on single GPU with 12G memory. 

\subsection{Supplementary Materials for the ImageNet Experiment}
\label{sec:supp_imagenet}
\begin{figure}[t]
\centering
    \begin{subfigure}{1\textwidth}
		\centering
		\includegraphics[width=1\linewidth]{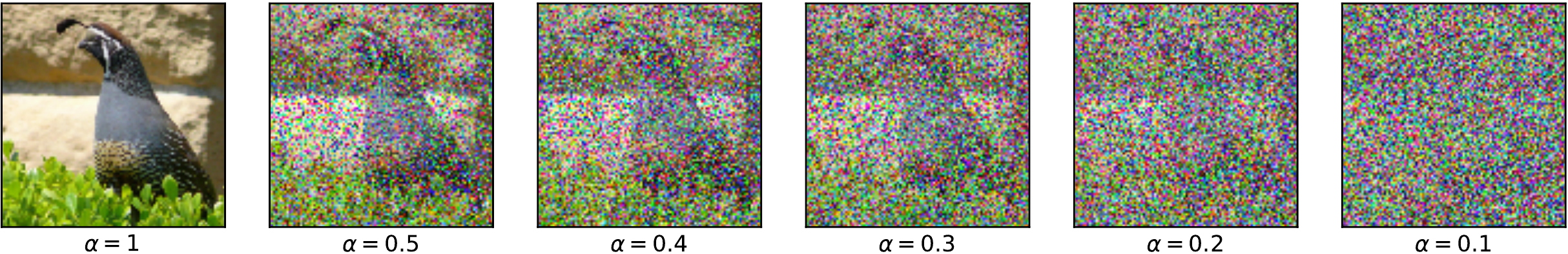}
	\end{subfigure}
    \vspace*{0.1in}
    \begin{subfigure}{1\textwidth}
		\centering
		\includegraphics[width=1\linewidth]{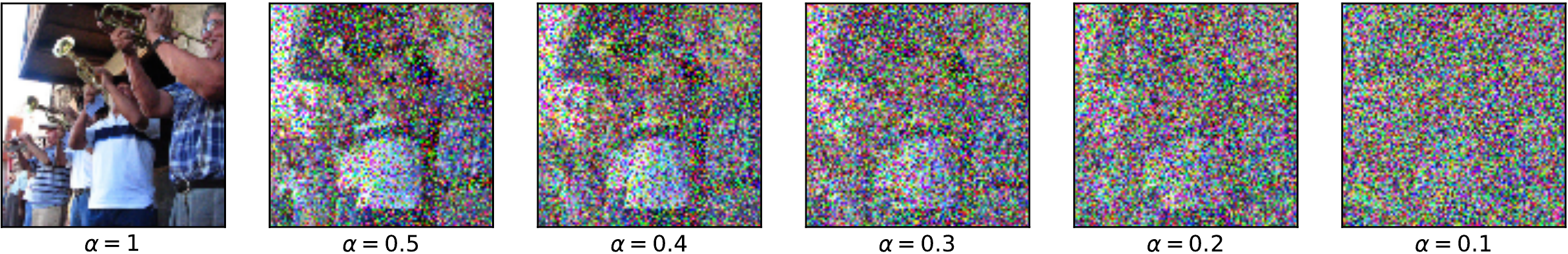}
	\end{subfigure}
	\vspace{-0.1in}
	\caption{ImageNet images corrupted by varying levels of the discrete noise.}
	\label{fig:imagenet_illustration}
\end{figure}

We use the \texttt{PyTorch}~\cite{paszke2017automatic} implementation provided by Cohen {et al.} \cite{cohen2019certified} as the backbone and implement our algorithm based on their pipeline. Thus, the training details are consistent to the one reported in their paper except that we use a different distribution. Here, we summarize some important details.  ResNet-50 is used as the base classifier for our ImageNet experiment, whose architecture is provided in \texttt{torchvision}.  After the randomization is done, we normalize each image by subtracting the dataset mean (0.485, 0.456, 0.406) and dividing by the standard deviation (0.229, 0.224, 0.225).  Parameters are optimized by SGD with momentum set as 0.9.  The learning rate is initially set to be 0.1 and annealed by a factor of 10 for every 30 epochs of training.  The total number of training epochs is 90. 
The batch size is $300$, parallelized across $2$ GPUs. 
We tune $\alpha \in \{0.1, 0.2, 0.3, 0.4, 0.5\}$ for the discrete distribution, and measure the performance in ACC@$r$, compared to the classifier under an additive isotropic Gaussian noise~\cite{cohen2019certified}.\footnote{We run their released model from \url{https://github.com/locuslab/smoothing}.}
The samples of the randomized image for each $\alpha$ are visualized in Figure~\ref{fig:imagenet_illustration}. We follow the prior work~\cite{cohen2019certified} to evaluate every 100th image in the validation set. The detailed accuracy numbers of our approach under different $\alpha$ and $r$ are available in Table~\ref{tab:detail:imagenet}. 

\begin{table*}[t]
  \caption{The guaranteed accuracy for different $\alpha$ of ResNet50 models smoothed by the discrete distribution on ImageNet. }\label{tab:detail:imagenet}
  \centering
  \begin{tabular}{cccccccccccccccccccc}
    \toprule
    \multirow{2}{*}{ $\alpha$ value} & \multicolumn{8}{c}{\small ACC@$r$}\\
    \cmidrule(lr{0pt}){2-8} 
    & \small $r=0$ & \small $r=1$ & \small $r=2$ & \small $r=3$ & \small $r=4$ & \small $r=5$ & \small $r=6$ & \small $r=7$\\
    \midrule
    \small $0.5$ & \small 0.666 & \small 0.412 & \small 0.388 & \small 0.000 & \small 0.000 & \small 0.000	& \small 0.000 & \small 0.000\\
    \small $0.4$ & \small 0.650 & \small 0.538 & \small 0.356 & \small 0.338 & \small 0.000 & \small 0.000	& \small 0.000 & \small 0.000\\
    \small $0.3$ & \small 0.592 & \small 0.516 & \small 0.314 & \small 0.300 & \small 0.274 & \small 0.234	& \small 0.000 & \small 0.000\\
    \small $0.2$ & \small 0.524 & \small 0.448 & \small 0.394 & \small 0.270 & \small 0.218 & \small 0.212	& \small 0.190 & \small 0.176\\
    \small $0.1$ & \small 0.350 & \small 0.314 & \small 0.282 & \small 0.248 & \small 0.212 & \small 0.182	& \small 0.150 & \small 0.100\\
    \bottomrule
  \end{tabular}
\end{table*}

\subsection{Supplementary Materials for the Chemical Property Prediction Experiment}
\label{sec:supp_chemical}
The dataset contains $1,513$ molecules (data points). 
We split the data into the training, validation, and testing sets with the ratio 0.8, 0,1, and 0,1, respectively. 
Following common practice in chemical property prediction~\cite{wu2018moleculenet}, the splitting is done based on the Bemis-Murcko scaffold~\cite{bemis1996properties}; 
the molecules within a split are inside different scaffolds from the other splits. We refer the details of scaffold splitting to \cite{wu2018moleculenet}. We observe similar experiment results when we use a random split. 

For both the decision tree and the randomly smoothed decision tree, we tune the depth limit in $\{6, 7, 8, 9, 10\}$. For the randomly smoothed decision tree, we also tune $\alpha \in \{0.7, 0.75, 0.8, 0.85, 0.9\}$.
The tuned $\alpha$ is $0.8$, and the tuned depth limits are $10$ for both models. 

\subsection{Computing Adversarial AUC}
\label{sec:qp}
Assume that there are $n+m$ data points, $n$ of them are positive instances, denoted as $A\triangleq \{x_1, \cdots, x_n\}$, and $m$ of them are negative instances, denoted $B\triangleq \{x_{n+1}, \cdots, x_{n+m}\}$. Denote the whole dataset as $X\triangleq A\cup B$. 
For data point $x\in X$, 
we may adversarially perturb $x$ up to the perturbation radius $r$, denoted as $x^r$.
Note that, since $\gY$ is binary, maximizing the probability for predicting one class can be equivalently done by minimizing the probability for predicting the other class.
Hence, we may use Algorithm~\ref{alg:dp} to find the adversaries for both the positive and negative examples.
Below we use the prediction probability for the class $1$ as the score. 
Denote the score of $x$ and $x^r$ as $s(x)$ and $s(x^r)$. 
For $x\in A$, we know that $s(x)\geq s(x^r)$, and for $x\in B$, $s(x)\leq s(x^r)$. 


If we are only allowed to perturb $k<n+m$ data points, 
to minimize AUC,
we aim to solve the following program: 

\begin{equation*}
\begin{array}{ll@{}ll}
\text{minimize}  & \displaystyle\sum\limits_{i=1}^{n}
\displaystyle\sum\limits_{j=1}^{m}
\Big[a_i b_j 
\hat \one(s(x_i^r),s(x_{j+n}^r)) 
+
a_i (1-b_j)
\hat \one(s(x_i^r),s(x_{j+n})) \\
& +
(1-a_i) b_j
\hat \one(s(x_i),s(x_{j+n}^r)) 
+
(1-a_i) (1-b_j)
\hat \one(s(x_i),s(x_{j+n})) 
\Big]\\
\text{subject to}& \displaystyle\sum\limits_{i\in[n]}   a_i+ \sum\limits_{j\in [m]} b_j \leq k,  \\
 &      a_{i} \in \{0,1\}, ~~~i=1 ,..., n\\
 &      b_{j} \in \{0,1\}, ~~~j=1 ,..., m
\end{array}
\end{equation*}

We may use standard mixed-integer programming solvers like Gurobi to solve the program. 
Here we use $a_i$ to denote whether data point $x_i\in A$ is perturbed, and $b_j$ to denote whether  data point $x_{j+n}\in B$ is perturbed. The function $\hat \one(x, x')$ is an indicator function defined as 
\begin{align}
    \hat \one(x, x')
     \triangleq \begin{cases}
         1,  &\text{if }x>x',\\
        0.5&\text{if } x=x',\\
         0, &\text{if }x<x'.
        \end{cases} 
\end{align}


\end{document}